\documentclass[letterpaper, 10 pt, conference]{ieeeconf}  %

\IEEEoverridecommandlockouts                              %

\let\labelindent\relax

\usepackage[para]{footmisc}
\usepackage{wrapfig}
\usepackage{capt-of}
\usepackage[pagebackref=true,breaklinks=true,colorlinks=true,bookmarks=false,citecolor=blue]{hyperref}

\usepackage{graphics,enumitem} %
\usepackage{epsfig} %
\usepackage{mathptmx} %
\usepackage{tabularx}
\usepackage{times} %
\usepackage{amsmath} %
\usepackage{multicol}
\usepackage{amssymb}  %
\usepackage{booktabs}
\usepackage{multirow}
\usepackage{algpseudocode}
\usepackage{algorithm}
\usepackage{gensymb}
\usepackage{amssymb}
\usepackage{makecell}
\usepackage{caption}
\usepackage{subcaption}
\usepackage{url}
\usepackage{multicol}
\usepackage{wrapfig,lipsum,booktabs}
\usepackage{xcolor,colortbl}
\usepackage{color}
\usepackage{cite}
\usepackage{balance}
\usepackage{physics}
\usepackage{float}
\usepackage{listings}
\usepackage{xspace}
\usepackage{bbm}
\usepackage{longtable}
\usepackage[pagebackref=true,breaklinks=true,colorlinks=true,bookmarks=false,citecolor=blue]{hyperref}
\hypersetup{
colorlinks=true,
linkcolor=blue,
filecolor=magenta,      
urlcolor=blue,
citecolor=blue
}
\newcommand{\ours}{Neural MP\xspace}
\definecolor{Gray}{gray}{0.9}
\title{\bf
Neural MP: A Generalist Neural Motion Planner
}
\vspace{-10pt}
\author{Murtaza Dalal$^{\ast}$\qquad Jiahui Yang$^{\ast}$\qquad Russell Mendonca \\ Youssef Khaky\qquad Ruslan Salakhutdinov\qquad Deepak Pathak\\
\small Carnegie Mellon University
}

\begin{document}
\makeatletter
\let\@oldmaketitle\@maketitle%
\renewcommand{\@maketitle}{\@oldmaketitle%
\includegraphics[width=1\linewidth]{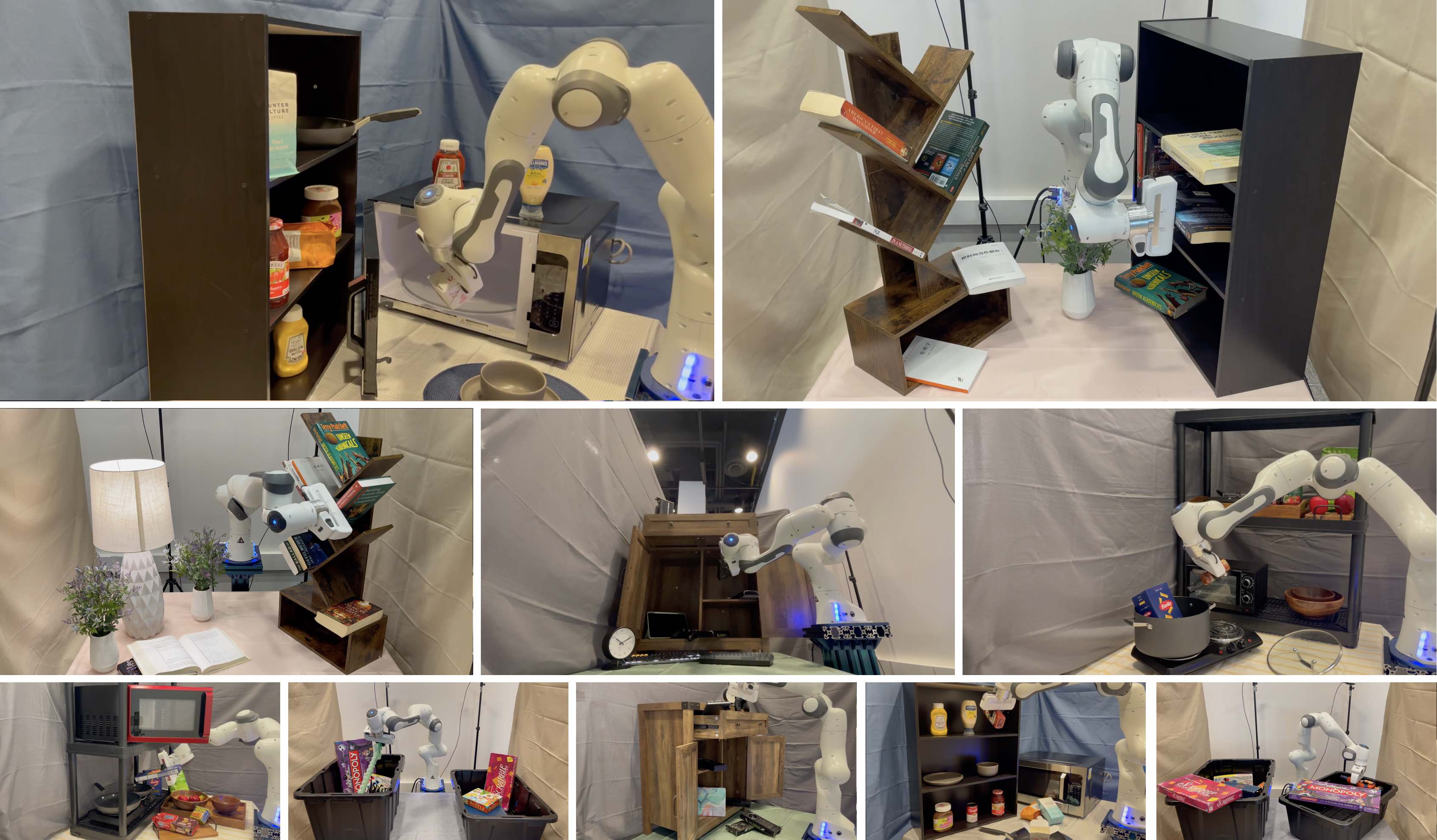}
  \centering
  \vspace{-15pt}
  \captionof{figure}{\small \textbf{Neural Motion Planning at Scale in the Real World} Our approach enables a \textit{single}, \textit{generalist} neural network policy to solve motion planning problems across diverse setups; Neural MP can generate collision free motions for a wide array of unseen tasks significantly faster and with higher success than traditional as well as learning-based motion planning approaches.}
  \label{fig:teaser}
  }
\makeatother
\maketitle
\thispagestyle{empty}
\pagestyle{empty}

\begin{abstract} 
The current paradigm for motion planning generates solutions from scratch for every new problem, which consumes significant amounts of time and computational resources. 
For complex, cluttered scenes, motion planning approaches can often take minutes to produce a solution, while humans are able to accurately and safely reach any goal in seconds by leveraging their prior experience. 
We seek to do the same by applying data-driven learning at scale to the problem of motion planning.
Our approach builds a large number of complex scenes in simulation, collects expert data from a motion planner, then \emph{distills} it into a reactive generalist policy. 
We then combine this with lightweight optimization to obtain a safe path for real world deployment. 
We perform a thorough evaluation of our method on 64 motion planning tasks across four diverse environments with randomized poses, scenes and obstacles, in the real world, demonstrating an improvement of 23\%, 17\% and 79\% motion planning success rate over state of the art sampling, optimization and learning based planning methods. 
Video results available at \texttt{\href{https://mihdalal.github.io/neuralmotionplanner}{mihdalal.github.io/neuralmotionplanner}}. 
\end{abstract}

\section{Introduction}

Motion planning is a longstanding problem of interest in robotics, with
previous approaches ranging from potential fields~\cite{khatib1986real,warren1989global,quinlan1993elastic}, sampling (RRTs and Roadmaps)~\cite{kavraki1996probabilistic,lavalle2001rapidly,lazyprm,karaman2011sampling,kuffner2000rrt,bit*,strub2020adaptively}, search (A*)~\cite{hart1968formal,likhachev2003ara,koenig2006new} and trajectory optimization~\cite{ratliff2009chomp,schulman2014motion,dragan2011manipulation,sundaralingam2023curobo}. 
Despite being ubiquitous, these methods are often slow at producing solutions since they 
largely plan from scratch at test time, re-using little to no information outside of the current problem and what is engineered by a human designer. Since motion-planning is a core component of the robotics stack for manipulation, its speed, capability and ease of use form a core bottleneck to developing efficient and reliable manipulation systems.
\let\thefootnote\relax\footnote{\textsuperscript{*}equal contribution.}

On the other hand, humans can generate motions in a closed loop manner, move quickly, react to various dynamic obstacles, and generalize across a wide distribution of problem instances. 
Rather than planning open loop from scratch, people draw on their vast amounts of experience moving and interacting with their environment while reactively adjusting their movements in order to quickly and efficiently move about the world. How can we create motion planners with similar properties? 
In this work, we argue that \textit{distillation} at scale is the answer: we can \textit{distill} the planning process into a \textbf{reactive, generalist} neural policy.

The primary challenge in training data-driven motion planning is the data collection itself, as scaling robotic data collection in real-world requires significant human time and effort. Recently, there has been a concerted effort to scale up data collection for robot tasks~\cite{padalkar2023open, khazatsky2024droid}. However, the level of diversity of scenes and arrangement of objects is still limited, especially for learning obstacle avoidance behavior that scales to the real world. Constructing such setups with diverse obstacle arrangements with numerous objects is prohibitively expensive in terms of cost and labor.

Instead, we leverage simulation, which makes it cheap and easy to obtain diverse data, is highly scalable via parallelization, and runs significantly faster than real world. Recent approaches have shown great promise in enabling policy learning for high-dof robots~\cite{lee2020learning, zhuang2023robot, cheng2023extreme, kumar2021rma, haarnoja2024learning,agarwal2023legged}. We build a large number of complex environments by combining procedural, programmatic assets with models of everyday objects sampled from large 3D datasets. These are used to collect expert data from state-of-the-art (SOTA) motion planners~\cite{strub2020adaptively}, which we then \emph{distill} into a reactive, generalist policy. Since this policy has seen data from \textbf{1 million} scenes, it is capable of generalizing to novel obstacles and scene configurations that it has never seen before. 
However, deploying neural policies in the real world might be unsafe for the system due to the potential of collisions. We mitigate this by using a linear model to predict future states the robot will end up in and run optimization to ensure a safe path.

Our core contribution is a SOTA motion planner that runs zero-shot on any environment, with more accuracy and in orders of magnitude less execution time. We demonstrate that large scale data generation in simulation can enable training generalist policies that can be successfully deployed for real-world motion planning tasks.
To our knowledge, \ours is the first work to demonstrate that such a neural policy can generalize to a broad set of out-of-distribution of real-world environments, generalizing across tasks with significant variation across poses, objects, obstacles, backgrounds, scene arrangements, in-hand objects, and start/goal pairs. 
Specifically, we propose a simple, scalable approach for training and deploying fast, general purpose neural motion planners: 1) \textbf{large-scale procedural scene generation} with diverse environments in realistic configurations, 2) \textbf{multi-modal sequence modeling} for fitting to sampling-based motion planning data and 3) \textbf{lightweight test-time optimization} to ensure fast, safe, and reliable deployment in the real world. 
We execute a thorough real-world empirical study of motion-planning methods, evaluating our approach on \textbf{64} real world motion planning tasks across \textbf{four} diverse environments, demonstrating a motion planning success rate improvements of \textbf{23\%} over sampling-based, \textbf{17\%} over optimization-based and \textbf{79\%} over neural motion planning methods.

\setcounter{figure}{1}
\begin{figure*}[t]
    \centering
    \includegraphics[width=\textwidth]{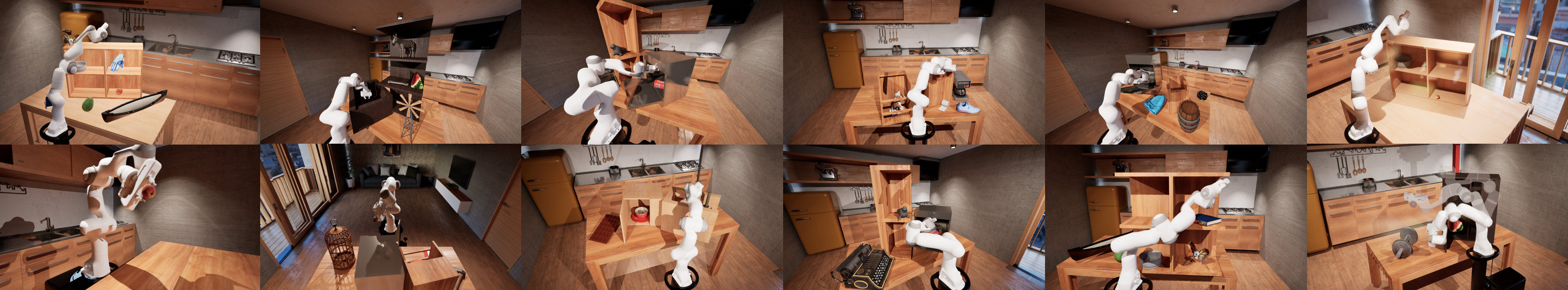} 
     \vspace{-15pt}
    \caption{\small \textbf{Visualization of Diverse Simulation Training Environments}: We train \ours on a wide array of motion planning problems generated in simulation, with significant pose, procedural asset, and mesh configuration randomization to enable generalization. }
    \label{fig:sim vis}
\end{figure*}

\section{Related Work}

\noindent \textbf{Approaches for Training General-Purpose Robot Policies}
Prior work on large scale imitation learning using expert demonstrations~\cite{brohan2022rt,brohan2023rt,padalkar2023open,dalal2023optimus,shridhar2023perceiver,khazatsky2024droid} has shown that large models trained on large datasets can demonstrate strong performance on challenging tasks and some varying levels of generalization. On the other hand, sim2real transfer of RL policies trained with procedural scene generation has demonstrated strong capabilities for producing generalist robot policies in the locomotion regime~\cite{kumar2021rma,zhuang2023robot,agarwal2023legged,cheng2023extreme}. 
In this work, we combine the strengths of these two approaches to produce powerful neural motion planning policies. We {propose} a method for procedural scene generation in simulation {and combine it with} large scale imitation learning to produce strong priors which we transfer directly to over 64 motion planning problems in the real world.

\noindent \textbf{Procedural Scene Generation for robotics}
Automatic scene generation and synthesis has been explored in vision and graphics~\cite{wang2021sceneformer,chang2015text,chang2014learning,ritchie2019fast} while more recent work has focused on embodied AI and robotics settings~\cite{deitke2022️,wang2023robogen,katara2023gen2sim,dalal2023optimus}.
In particular, methods such as Robogen~\cite{wang2023robogen} and Gen2sim~\cite{katara2023gen2sim} use LLMs to propose tasks and build scenes using existing 3D model datasets~\cite{deitke2023objaverse} or text-to-3D~\cite{poole2022dreamfusion,wang2023score} and then decompose the tasks into components for RL, motion-planning and trajectory optimization to solve in simulation. Our method is instead rule-based rather than LLM-based, is designed specifically for generating data to train neural motion planners (see Sec.~\ref{subsec:datagen}), and demonstrates that policies trained on its data can indeed be transferred to the real world. MotionBenchmaker~\cite{chamzas2021motionbenchmaker}, on the other hand, is similar to our data generation method in that it autonomously generates scenes using programmatic assets. However, the datasets generated by MotionBenchmaker are not realistic: floating robots, a single major obstacle per scene and primitive objects that are spaced far apart. By comparison, the scenes and data generated by our work (Fig.~\ref{fig:sim vis}) are considerably more diverse, containing additional programmatic assets that incorporate articulations (microwave, dishwasher), multiple large obstacles per scene (up to 5), complex meshes sampled from Objaverse~\cite{deitke2023objaverse}, and tightly packed obstacles.

\noindent \textbf{Neural Motion Planning}
Finally, there is a large body of recent work~\cite{qureshi2019motion,fishman2023motion,qureshi2020neural,carvalho2023motion,ichter2020broadly,saha2023edmp,huang2023diffusion} focused on imitating motion planners in order to accelerate planning. MPNet~\cite{qureshi2019motion,johnson2020dynamically,qureshi2020neural} trains a network to imitate motion planners, then integrates this prior into a search procedure at test time. Our method leverages large scale scene generation and sequence modeling, enabling it to use a faster optimization process at test time while obtaining strong results across a diverse set of tasks. 
M$\pi$Nets~\cite{fishman2023motion} trains the SOTA neural motion planning policy using procedural scene generation and demonstrates transfer to the real world. 
Our approach is similar, albeit with 1) much more diverse scenes via programmatic asset generation and complex real-world meshes, 2) a more powerful learning architecture and multi-modal output distributions and 3) test-time optimization to improve performance at deployment, enabling significantly improved performance over M$\pi$Nets.

\section{Neural Motion Planning}
Our approach enables generalist neural motion planners, by leveraging large amounts of training data generated in simulation via expert planners. The policies can generalize to out-of-distribution settings by using powerful deep learning architectures along with diverse, large-scale training data. To further improve the performance of these policies at deployment, we leverage test time optimization to select the best path out of a number of options. We now describe each of these pieces in more detail. 

\subsection{Large-scale Data Generation}
\label{subsec:datagen}
One of the core lessons of the deep learning era is that the quality and quantity of data is crucial to train broadly capable models. We leverage simulation to generate vast datasets for training robot policies. Our approach generates assets using programmatic generation of primitives and by sampling from diverse meshes of common objects. These assets are combined to create complex scenes resembling real world scenarios (Fig.~\ref{fig:sim vis}), as described in Alg.~\ref{alg:scene_gen}.

\begin{figure*}[t]
    \centering
    \includegraphics[width=\textwidth]{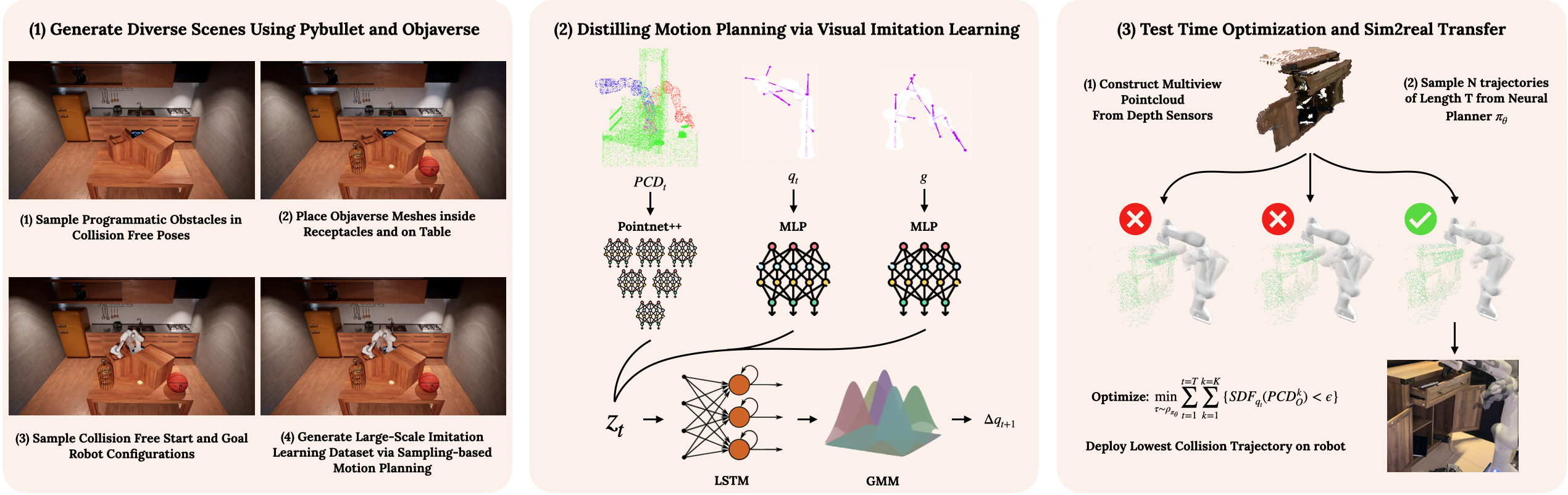} 
    \caption{\small \textbf{Method Overview}: We present Neural Motion Planners, which consists of 3 main components. \textbf{Left}: Large Scale data generation in simulation using expert planners \textbf{Middle}: Training deep network models to perform fast reactive motion planning \textbf{Right}: Test-time optimization at inference time to improve performance. }
    \label{fig:method}
\end{figure*}

\noindent \textbf{Procedural Generation From Primitives}
How do we generate a large enough number of diverse environments to train a generalist policy? Hand designing each environment is tedious, requiring significant human effort per scene, which doesn't scale well. Instead, we take the approach of \emph{procedural} scene generation, using a set of six \emph{parametrically variable} categories - shelves, cubbies, microwaves, dishwashers, open boxes, and cabinets. These categories are representative of a large set of objects in everyday scenarios that robots encounter and have to avoid colliding with. Each category instance is constructed using a combination of primitive cuboid objects and is parameterized by category specific parameters which define the asset. 
Specifically a category instance $g$ is comprised of N cuboids $g = \{ x_0..x_i...x_N\}$, which satisfy the category level constraint  given by $C(g)$. For controlled variation within each category, we make use of \emph{parametric} category specific generation functions
 $X(\textbf{p}) = \{ x_0..x_i.x_N\}, \text{s.t. } C(X(\textbf{p}))$, where $p$ specifies the size and scale of each of the cuboids, their relative positions, and specific axes of articulation. The constraint C(.) relates to the relative positions, scales and orientations of the different cuboids, \textit{e.g} for the microwave category the constraint ensures each of the walls are of the same height, and that the microwave has a hinge door.

\begin{algorithm}

\caption{Procedural Scene Generation}
\label{alg:scene_gen}
\begin{algorithmic}[1]
\Require Asset category generators $ \{ X_i(\textbf{p}) \}_{0,1..G}$  
\Require Number of scenes $N$
\Require Max objects per scene $K$
\Require Collision checker $Q$
\For {scene 1: N}
    \State Initialize scene $S = \{ \}$
    \State Sample number of assets $k \sim [1,...K]$
    \For {asset 1:k}
        \State Sample asset category $g \sim [0,..N]$
        \State Sample asset parameter $p$
        \State Sample asset $x \sim X_g(p)$
        \While {$Q(S, x)$}
            \For {each asset $s_i$ in $S$}
                \State $n_i$ = collision normal b/n $x$ and $s_i$
            \EndFor
            \State Effective collision normal $n = \sum n_i$
            \State Update $p$ so $X_g(p)$ center is shifted along $n$
        \EndWhile
        \State Add asset $x$ to scene $S$
    \EndFor
    \State yield scene S
\EndFor
\end{algorithmic}
\end{algorithm}

\noindent \textbf{Objaverse Assets For Everyday Objects} While programmatic generation can create a large number of scenes using the defined categories, there are a large number of everyday objects the robot might encounter that lie outside this distribution. For example, a robot will need to avoid collisions with potted plants, bowls and utensils while moving between locations, as shown in Fig~\ref{fig:teaser}.
To better handle these settings, we augment our dataset with objects sampled from the recently proposed large-scale 3D object dataset, Objaverse~\cite{deitke2023objaverse}. This dataset contains a wide variety of objects that the neural planner is likely to observe during deployment, such as comic books, jars, record players, caps, etc. We sample these Objaverse assets in the task-relevant sampling location of the programmatic asset(s) in the scene, such as between shelf rungs, inside cubbies or within cabinets.

\noindent \textbf{Complex Scene Generation}
The scenes we use comprise combinations of the procedurally generated assets built from primitives, and the Objaverse assets arranged on a flat tabletop surface. 
A naive approach to constructing realistic scenes is to use rejection-sampling based on collision. This involves iteratively sampling assets on a surface, and re-sampling those that collide with the current environment. However, as the number, size and type of objects increases, so does the probability of sampling assets that are in collision, making such a process prohibitively expensive to produce a valid configuration. In addition, this is biased towards simple scenes with few assets that are less likely to collide, which is not ideal for training generalist policies. Instead, we propose an approach that iteratively adds assets to a scene by adjusting their position using the effective collision normal vector, computed from the existing assets in the scene. Please see Alg.~\ref{alg:scene_gen} and the Appendix for additional details.

\noindent \textbf{Motion Planning Experts}: 
To collect expert data in the diverse generated scenes, we leverage SOTA sampling-based motion planners due to their (relative) speed as well as ease of application to a wide array of tasks. Specifically, we use Adaptively Informed Trees~\cite{strub2020adaptively} (AIT*), an almost-surely asymptotically optimal sampling-based planner to produce high-quality plans using privileged information, namely access to a perfect collision checker in simulation. 
How do we ensure that the planner is evaluated between points in the scene that require it to maneuver around obstacles? 
We generate tight-space configurations by sampling end-effector poses from specific locations (\textit{e.g.}, inside a cubby or microwave) and by using inverse kinematics (IK) to derive the joint pose. Tight-space configurations are sampled 50\% of the time, to ensure that we collect trajectories where the robot moves around obstacles, as opposed to taking straight line paths between nearby free space points. 
Additionally, we spawn objects grasped in the end-effectors, with significant randomization including boxes, cylinders, spheres or even Objaverse meshes.
Importantly, we found that naively imitating the output of the planner performs poorly in practice as the planner output is not well suited for learning. Specifically, plans produced by AIT* often result in way-points that are far apart, creating large action jumps and sparse data coverage, making it difficult for networks to fit the data. To address this issue, we perform smoothing using cubic spline interpolation while enforcing velocity and acceleration limits. We found that smoothing is crucial for learning performance as it ensures action size limits for each time-step transition.

\subsection{Generalist Neural Policies}
We would like to obtain agents that can use diverse sets of experiences to plan efficiently in new settings.
In order to build generalist neural motion planning policies, we need an observation space amenable to sim2real transfer, and utilize an architecture capable of absorbing vast amounts of data.

\noindent \textbf{Observations}: We begin by addressing the sim2real transfer problem, which requires considering the observation and action spaces of the trained policy. With regards to observation, point-clouds are a natural representation of the scene for transfer~\cite{fishman2023motion,christen2023handoversim2real,jiang2024transic,chen2023visual,wu2023sim2real}, as they are 3D points grounded in the base frame of the robot and therefore view agnostic, and largely consistent between sim and real. 
We include proprioceptive and goal information in the observations, consisting of the current joint angles $q_t$, the target joint angles $g$, in addition to the point-cloud $PCD$.

\noindent \textbf{Network Architecture}: We require an architecture capable of scaling with data while performing well on multi-modal sequential control problems, \textit{e.g.} motion planning. To that end, we design our policy $\pi$ (visualized in Fig.~\ref{fig:method}) to be a sequence model to imitate the expert using a notion of history which is useful for fitting privileged experts using partially observed data~\cite{dalal2023optimus}. In principle, any sequence modeling architecture could be used, but in this work, we opt for LSTMs for their fast inference time and comparable performance to Transformers on our datasets (see Appendix). We operate the LSTM policy over joint embeddings of $PCD_t$, $q_t$, and $g$ with a history length of 2. 
We encode point-clouds using PointNet++~\cite{qi2017pointnet++}, while we use MLPs to encode $q_t$ and $g_t$. We follow the design decisions from M$\pi$Nets regarding point-cloud observations: we segment the robot point-cloud, obstacle point-cloud and the target robot point-cloud before passing it to PointNet++. For each time-step, we concatenate the embeddings of each of the observations together into one vector and then pass them into the LSTM for action prediction.
For the output of the model, note that sampling-based motion planners such as AIT* are heavily multi-modal: for the same scene they may give entirely different plans for different runs. As a result, we require an expressive, multi-modal distribution to effectively capture such data, for which we use a Gaussian Mixture Model (GMM).
Specifically, \ours predicts a GMM distribution over delta joint angles ($\Delta q_{t+1}$), which are used to compute the next target joint way-point during deployment: $q_{t+1} = q_t + \Delta q_{t+1}$. 
As we show in our experiments, for fitting to sampling-based motion planning, minimizing the negative log-likelihood of the GMM outperforms the PointMatch loss from M$\pi$Nets, Diffusion~\cite{saha2023edmp} and Action-chunking~\cite{zhao2023learning} (Sec.~\ref{sec:result} and Appendix).

\subsection{Deploying Neural Motion Planners}
~\label{subsec:deployment}
\noindent \textbf{Test time Optimization} While our base neural policy is capable of solving a wide array of challenging motion planning problems, we would still like to ensure that these motions are safe to be deployed in real environments. We enable this property by combining our learned policy with a simple light-weight optimization procedure at inference time.
This relies on a simple model that 
assumes the obstacles do not move and the controller can accurately reach the target way-points. Given world state $s = [q, e]$ (e is the environment state), the predicted world state is $s' = [q + \hat{a}$, e] where $\hat{a}$ is the policy prediction. With this forward model, we can sample N trajectories from the policy using the initial scene point-cloud to provide the obstacle representation and estimate the number of scene points that intersect the robot
using the linear forward model. We then optimize for the path with the least robot-scene intersection in the environment, using the robot Signed Distance Function (SDF).
Specifically, we optimize the following objective at test time: 
\begin{equation}
    \min_{\tau \sim \rho_{\pi_{\theta}}} \sum_{t=1}^{t=T} \sum_{k=1}^{k=K} \mathbbm{1} \{ SDF_{q_t}(PCD_{O}^{k}) < \epsilon\}
\end{equation}
in which $\rho_{\pi_{\theta}}$ is the distribution of trajectories under policy $\pi_{\theta}$ with a linear model as described above, $PCD_{O}^{k}$ is the kth point of the obstacle point-cloud (with max $K=4096$ points) and $SDF_{q_t}$ is the SDF of the robot at the current joint angles. In practice, we optimize this objective with finite samples in a single step, computing the with minimal objective value by selecting the path with minimal objective value across 100 trajectories. We include a detailed analysis of the properties of our proposed test-time optimization approach in the Appendix.

\noindent \textbf{Sim2real and Deployment}
For executing our method on a real robot, we predict delta joint way-points which we then linearly interpolate and execute using a joint space controller. Our setup includes four extrinsically calibrated Intel RealSense cameras (two 435 and two 435i) positioned at the table's corners. 
To produce the segmented point cloud for input to the robot, we compute a point-cloud of the scene using the 4 cameras, segment out the partial robot cloud using a mesh-based representation of the robot to exclude points. 
We then generate the current robot and target robot point clouds using forward kinematics on the mesh-based representation of the robot and place them into the scene. For real-world vision-based collision checking, we calculate the SDF between the point cloud and the spherical representation of the robot, enabling fast SDF calculation (0.01-0.02s per query), though this method can lack precision for tight spaces.

\section{Experimental Setup}
\begin{figure}
    \centering
    \begin{subfigure}[b]{1.0\linewidth}
        \centering
        \includegraphics[width=1.\linewidth]{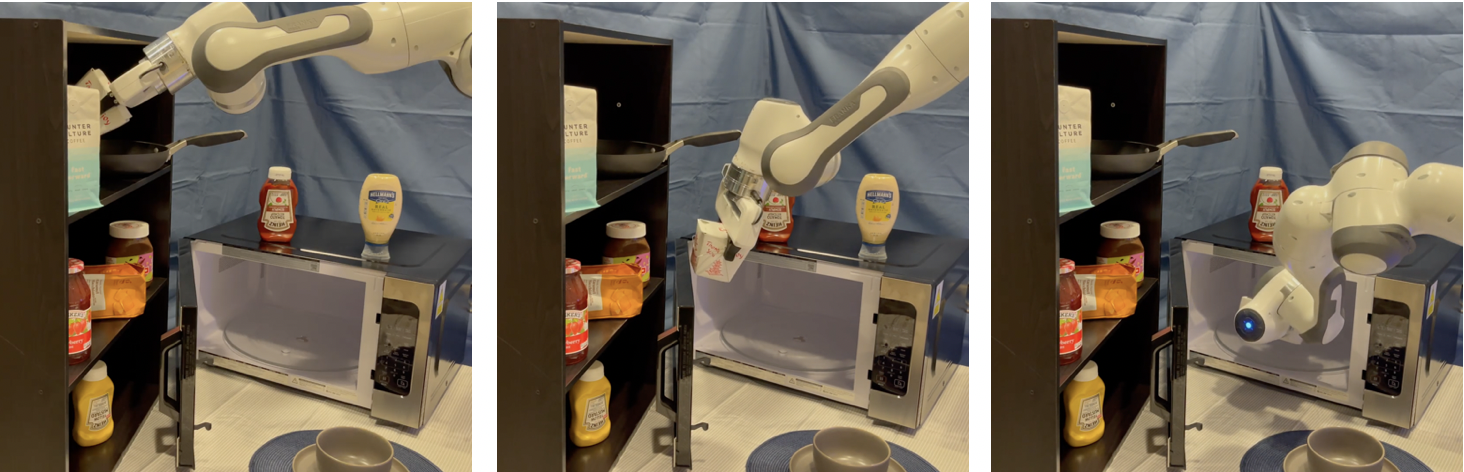}
        \vspace{-15pt}
        \caption{\small Sampling-based planners struggle with tight spaces, 
        a regime in which \ours performs well.
        }
        \label{fig:sub1}
    \end{subfigure}
    \hfill
    \begin{subfigure}[b]{1.0\linewidth}
        \centering
        \includegraphics[width=1.\linewidth]{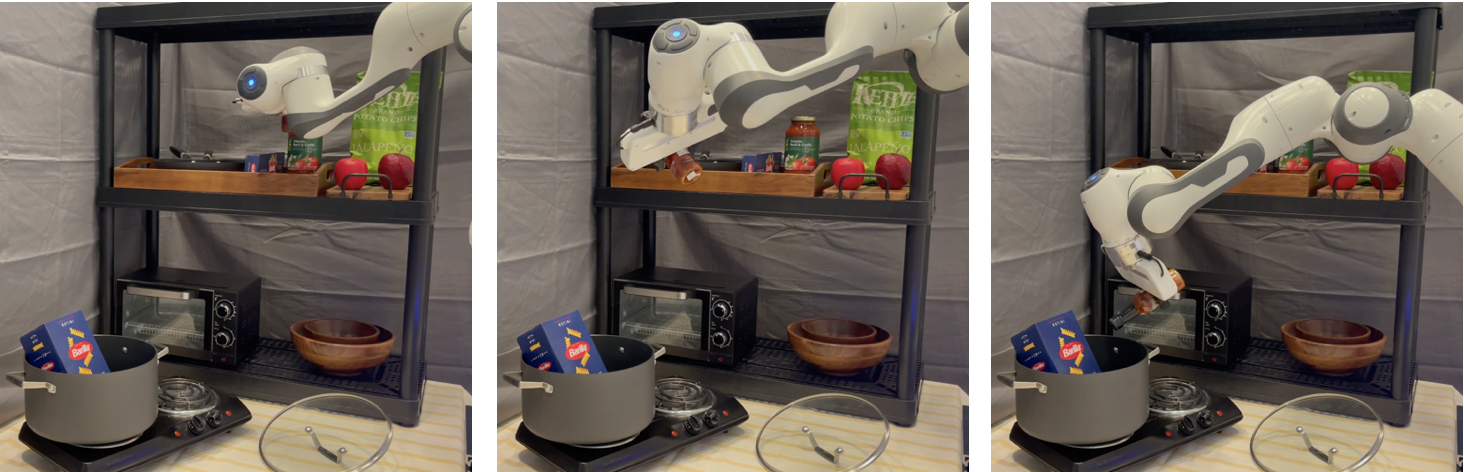}
        \vspace{-15pt}
        \caption{\small Our method is able to motion plan with objects in-hand, a crucial skill for manipulation.}
        \label{subfig: in hand rollout}
    \end{subfigure}
    \hfill
    \begin{subfigure}[b]{1.0\linewidth}
        \centering
        \includegraphics[width=1.\linewidth]{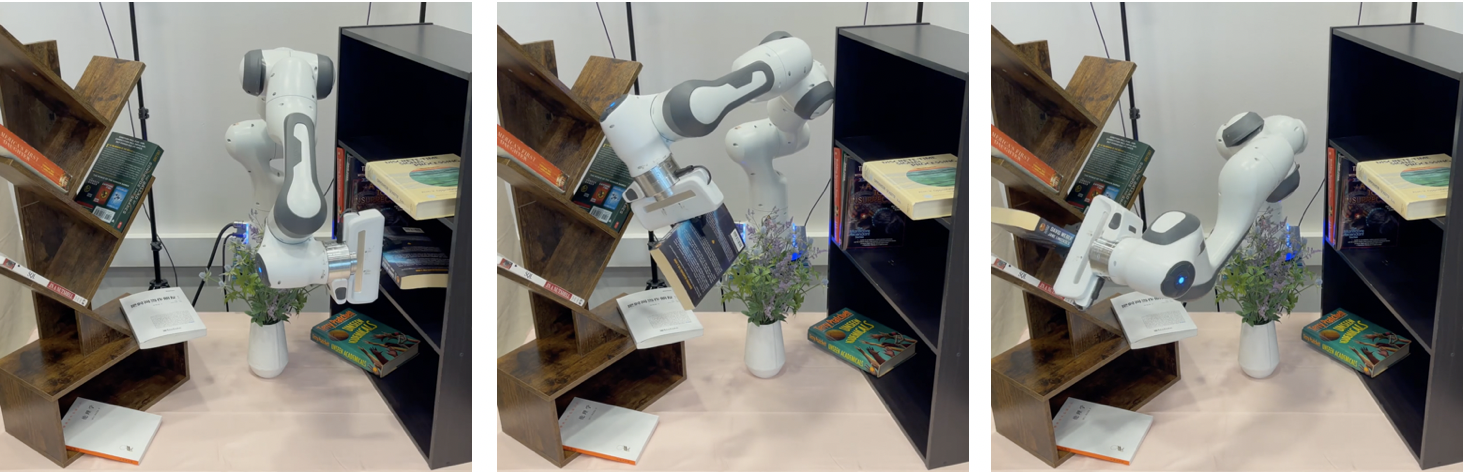}
        \vspace{-15pt}
        \caption{\small Our policy has not been trained on this bookcase, yet it is able to insert the book into the correct location.}
        \label{fig:bookcase rollout}
    \end{subfigure}
    \vspace{-15pt}
    \caption{\small Emergent Capabilities of \ours}
    \vspace{-15pt}
    \label{fig:rollouts}
\end{figure}
In our experiments, we consider motion planning in four different real world environments containing obstacles (see Appendix). Importantly, these are not included as part of the training set, and thus the policy needs to generalize to perform well on these settings. We begin by describing our environment design, then each of the environments, and finally our evaluation protocol and comparisons.

\noindent \textbf{Environment Design}
We evaluate our motion planner on tabletop motion planning tasks which we subdivide into \textit{environments}, \textit{scenes}, and \textit{configurations}. We evaluate on four different environments, with each environment containing 1-2 large receptacles that function as the primary obstacles. For each environment, we have four different scenes which involve significant pose variation (over the entire tabletop) of the primary obstacles, table height randomization, as well as randomized selection, pose and orientation of objects contained within the receptacles. For each environment, we have two scenes with obstacles and two without obstacles. For each scene, we evaluate on four different types of start ($q_0$) and goal ($g$) angle pairs: 1) free space to free space, 2) free space to tight space 3) tight space to free space 4) tight space to tight space. 
Free space configurations do not have an obstacle in the vicinity of the end-effector, while tight space configurations generally have obstacles on most sides of the end-effector. 
Our four environments are 1) \textbf{Bins}: moving in-between, around and inside two different industrial bins 2) \textbf{Shelf}: moving in-between and around the rungs of a black shelf 3) \textbf{Articulated}: moving inside and within cubbies, drawers and doors 4) \textbf{in-hand}: moving between rungs of a shelf while holding different objects. 

\noindent \textbf{Evaluation Protocol}
We evaluate all methods on open loop planning performance for fairness, though our method, just like M$\pi$Nets, is capable of executing trajectories in a closed loop manner. For neural planners such as our method and M$\pi$Nets, this involves generating an open loop path by passing the agent's predictions back into itself using a linear model for the next state, as described in Sec.~\ref{subsec:deployment}. We then execute the plans on the robot, recording the success rate of the robot in reaching the goal, its collision rate and the time taken to reach the goal. We follow M$\pi$Nets' definition of success rate: reaching within 1cm and 15 degrees of the goal end-effector pose of the target goal configuration while also not colliding with anything in the scene. In practice, our policy achieves orientation errors significantly below this threshold, 2 degrees or less.

\noindent \textbf{Comparisons}
We propose three baselines for real-world comparisons to evaluate different aspects of our method's capabilities. We compare against sampling-based motion planning, which is expensive to run but has strong guarantees on performance. The first baseline is the expert we use to train our model, AIT* with 80 seconds of planning time. We run this planner with the same vision-based collision checker used by our method in the real world.
AIT*-80s is impractical to deploy in most settings due to its significant planning time. Thus, we compare to a faster variant of AIT* with 10 seconds of planning time, which uses comparable time to our method (Note: AIT*-3s is unable to find a plan for any real world task). 
Next, we compare against Curobo~\cite{sundaralingam2023curobo}, a SOTA motion-generation method which performs GPU-parallelized optimization and is orders of magnitude faster than AIT*. We run this baseline with a voxel-based collision checker and optimize its voxel resolution per task due to its sensitivity to that parameter. Finally, we compare against the SOTA neural motion planning approach, M$\pi$Nets.

\begin{table}[]
    \centering
    \resizebox{\linewidth}{!}{%
        \begin{tabular}{lcccc}
        \toprule
         & Bins & Shelf & Articulated & Average \\
        
        \multicolumn{2}{l}{\textit{Sampling-based Planning}:}\vspace{0.4em}\\
        \texttt{AIT*-80s~\cite{strub2020adaptively}} & 93.75 & 75.0 & 50.0 & 72.92\\
        \texttt{AIT*-10s (fast)~\cite{strub2020adaptively}} & 75.0 & 37.5 & 25.0 & 45.83\\
        \multicolumn{2}{l}{\textit{Optimization-based Planning}:}\vspace{0.4em}\\
        \texttt{Curobo~\cite{sundaralingam2023curobo}} & 93.75 & 81.25 & 62.5 & 79.17\\
        \midrule
        \multicolumn{2}{l}{\textit{Neural}:}\vspace{0.4em}\\
        \texttt{M$\pi$Nets~\cite{fishman2023motion}} & 18.75 & 25.0 & 6.25 & 16.67  \\
        \midrule
        \texttt{\textbf{Ours}-Base Policy} & 81.25 &	75.0 &	43.75 & {66.67} \\
        \textbf{\texttt{Ours}} & \textbf{100} & \textbf{100} & \textbf{87.5} & \textbf{95.83}  \\ 
        \bottomrule
        \end{tabular}
        }
        \caption{\small 
        \ours performs best across each scene free-hand motion planning task, demonstrating greater improvement as the task complexity grows.}
        \label{tab: main results}
        \vspace{-10pt}
\end{table}
\section{Experimental Results}
\label{sec:result}
To guide our evaluation, we pose a set of experimental questions. 1) Can a single policy trained in simulation learn to solve complex motion planning tasks in the real world? 2) How does \ours compare to SOTA neural planning, sampling-based and trajectory optimization planning approaches? 3) How well does \ours extend to motion planning tasks with objects in-hand? 4) Can \ours perform dynamic obstacle avoidance? 5) What are the impacts key ingredients of \ours have on its performance? 

\noindent \textbf{Free Hand Motion Planning}
In this set of experiments, we evaluate motion planning the robot's hand is empty (Table~\ref{tab: main results}). We find that our base policy on its own performs comparably to AIT*-80s (66.67\% vs. 72.92\%) while only using 1s of planning time. When we include test-time optimization (3s of planning), we find that across all three tasks, \ours achieves the best performance with a 95.83\% success rate. In general, we find that Bins is the easiest task, with the sampling/optimization-based methods performing well, Shelf is a bit more difficult as it requires simultaneous vertical and horizontal collision avoidance, while Articulated is the most challenging task as it contains a diverse set of obstacles and tight spaces. \ours performs well across each task as it is trained with a diverse set of parametric objects that cover the types of real-world obstacles we encounter and it also incorporates complex meshes which cover the irregular geometries of the additional objects we include. 

In our experiments, M$\pi$Nets performs poorly across the board. We attribute this finding to 1) M$\pi$Nets is only trained on data in which the expert goes from tight spaces to tight spaces, which means the fails to generalize well to start/goal configurations in free space and 2) the end-effector point matching loss in M$\pi$Nets fails to distinguish between 0 and 180 degree rotations of the end-effector, so the network has not learned how to match ambiguous target end-effector poses. Note, even if we change the success rate metric for M$\pi$Nets to count 180 degree flipped end-effector poses as successes as well, the average success rate of M$\pi$Nets only improves from 16.67\% to 29.17\% - it is still far below the other methods. Meanwhile, failure cases for AIT* and Curobo are tight spaces for which vision-based collision checking is inaccurate and the probability of sampling/optimizing for a valid path is low. In contrast, our method performs well on each task, generalizing to 48 different unseen environment, scene, obstacle and joint configuration combinations.

\noindent \textbf{In-Hand Motion Planning}
In this experiment, we extend our evaluation to motion planning with objects in-hand, a crucial capability for manipulation. We evaluate \ours against running the neural policy without test time optimization and without including any Objaverse data, achieving 81\% performance vs. 31\% and 44\%. We visualize an example trajectory in Fig.~\ref{fig:rollouts}. Our method performs well on in-distribution objects such as the book and board game, but struggles on out of distribution objects such as the toy sword, which is double the size of objects at training time. 
We additionally deploy our method on significantly out of distribution objects such as the bookcase (Fig.~\ref{fig:bookcase rollout}) and find that \ours generalizes well to in-hand motion planning tasks such as inserting the book in the right rung. 

This experiment also serves as an ablation of our method, demonstrating the importance of test time optimization on out of distribution scenarios. For these tasks, the base policy performance results in a large number of collisions as two of the in-hand objects are out of distribution (sword and board game), but the optimization step is able to largely remove them and produce clean behavior that reaches the target without colliding. Additionally, this experiment demonstrates that the Objaverse data is crucial for the success of our method in the real world. Models trained only on cuboid-based parametric assets fail to generalize to the complexity of the real world (43.75\%) while those trained on Objaverse perform well (81.25\%), highlighting the importance of incorporating Objaverse meshes into scene generation.

\noindent \textbf{Dynamic Motion Planning}
In many real-world scenarios, the environment may be changing as the motion planner is acting.
We test how well \ours can motion plan in such settings by introducing obstacles into the environment while the motion planner is moving to a goal. We evaluate the motion planner on four different goals with three different added obstacles (drawer, monitor and pot). To handle dynamic obstacles, we run the neural motion planner closed loop and perform single-step test-time optimization. We compare against M$\pi$Nets and find that \ours performs $53\%$ better ($63.33\%$ vs. $10\%$), performing particularly well on the drawer and pot object while struggling on the monitor object which is significantly taller. We also qualitatively evaluate \ours on two significantly more challenging motion planning tasks in which we continuously move the obstacle into the robot's path and demonstrate that it can adjust its behavior to avoid collisions while reaching the goal.

\begin{table}
    \vspace{-10pt}
    \centering
    \resizebox{.9\linewidth}{!}{%
        \begin{tabular}{lcccc}
        \toprule
         & Global & Hybrid & Both & Average \\
        \midrule
        \multicolumn{2}{l}{\texttt{MPNet~\cite{qureshi2019motion}}}\vspace{0.4em}\\
        \textit{Hybrid Expert} & 41.33 & 65.28 & 67.67 & 58.09\\
        \multicolumn{2}{l}{\texttt{M$\pi$Nets~\cite{fishman2023motion}}}\vspace{0.4em}\\
        \textit{Global Expert} & 75.06 & 80.39 & 82.78 & 79.41\\
        \textit{Hybrid Expert} & 75.78 & 95.33 & 95.06 & 88.72\\
        \multicolumn{2}{l}{\texttt{EDMP~\cite{saha2023edmp}}}\vspace{0.4em}\\
        \textit{Global Expert} & 71.67 & 82.84 & 82.79 & 79.10\\
        \textit{Hybrid Expert} & 75.93 & 86.13 & 85.06 & 82.37\\
        \midrule
        \multicolumn{2}{l}{\texttt{Ours}}\vspace{0.4em}\\
        \textit{Global Expert} & \textbf{77.93} & 85.50 & 87.67 & 83.70\\
        \textit{Hybrid Expert} & 76.33 & \textbf{97.28} & \textbf{96.78} & \textbf{90.13}\\
        \bottomrule
        \end{tabular}
    }
    \caption{\small Performance comparison of neural motion planning methods across 5400 test problems in the M$\pi$Nets dataset in simulation. \ours achieves the SOTA results on these tasks.}
    \label{tab:sim results}
    \vspace{-10pt}
\end{table}
\noindent \textbf{Comparisons to Learning-based Motion Planners}
We next evaluate how \ours compares to two additional learning-based methods, MPNets~\cite{qureshi2019motion} and EDMP~\cite{saha2023edmp} (a Diffusion-based neural motion planner) as well as M$\pi$Nets~\cite{fishman2023motion} in simulation. We compare these three neural motion planning methods in simulation trained on the same dataset (from M$\pi$Nets) of 3.27 million trajectories. We train policies on the Global expert data and the Hybrid datasets and then evaluate on 5400 test problems across the Global, Hybrid and Both solvable subsets. We include numerical results Tab.~\ref{tab:sim results}, with numbers for the baselines taken from the EDMP and M$\pi$Nets papers. We find that across the board, Neural MP is the best learning-based motion planning method, outperforming both EDMP and M$\pi$Nets on the test tasks provided in the M$\pi$Nets paper. We attribute this to the use of sequence modelling, the ability of the GMM to fit multimodal data and test-time optimization to prune out any collisions.

\noindent \textbf{Data Scaling}
In order to understand the scaling of our method with data we evaluate how performance changes with dataset size. To do so, we train models with 1K trajectories, 10K trajectories and 100K trajectories. In these experiments, we train with subsets of our overall dataset and evaluate on held out simulation environments which are not sampled from the training distribution. 
While performance with a thousand trajectories is weak (15\%), we find rapid improvement as we increase the orders of magnitude of data (10K - 50\%, 100K - 65\%), with the model trained on 1M trajectories achieving 80\% success rate on entirely held out tight-space shelf and bin configurations, demonstrating that our method scales and improves with data. 

\noindent \textbf{Ablations}
We run ablations of components of our method (training objective, observation composition) in simulation to evaluate which have the most impact. For each ablation we evaluate performance on held out scenes. For training objective, we find that GMM (ours) outperforms L2 loss, L1 loss, and PointMatch Loss (M$\pi$Nets) by (7\%, 12\%, and 24\%) respectively. We find that including both $q$ and $g$ vectors is crucial for performance as we observe a 62\%, 65\%, and 75\% performance drop when using only $g$, only $q$ and neither $q$ nor $g$ respectively. 
We refer the reader to the Appendix for further analysis, discussion and results. 
\vspace{-0.1in}
\section{Discussion and Limitations}
\label{sec:conclusion}
In this work, we present \ours, a method that builds a data-driven policy for motion planning by scaling procedural scene generation, distilling sampling-based motion planning and improving at test-time via refinement. Our model demonstrably improves over the sampling-based planning in the real world, operating 2.5x-20x faster than AIT* while improving by over 20\% in terms of motion planning success rate. Notably, our model generalizes to a wide distribution of task instances and demonstrates favorable scaling properties. At the same time, there is significant room for future work to improve upon, our model 1) is susceptible to point-cloud quality, which may require improving 3D representations via implicit models such as NeRFs~\cite{mildenhall2021nerf}, 2) does not still handle tight spaces well, a capability which could be potentially acquired via RL fine-tuning of the base policy and 3) is slower than simply running the policy directly due to test-time optimization, which can be addressed by leveraging learned collision checking~\cite{murali2023cabinet,danielczuk2021object}.

\section{Acknowledgment}
We thank Shikhar Bahl, Ananye Agarwal, and Mihir Prabhudesai for their insightful discussions and feedback. We additionally thank Kenny Shaw, Rishi Veerapaneni, Ananye Agarwal, and Shikhar Bahl for feedback on early drafts of this paper. This work was supported in part by the NSF Graduate Fellowship, Apple, ONR MURI N00014-22-1-2773 and AFOSR FA9550-23-1-0747.

\bibliographystyle{IEEEtran}
\bibliography{main}

\clearpage

\section*{Appendix}

\section{Additional Real World Results and Analysis}
\label{app:analysis}

\subsection{Detailed Free Hand Motion Planning Results}
In this section we perform additional analysis of the free hand motion planning results from the main paper. We include a more detailed version of the main result table (Tab.~\ref{tab:main detailed results}). In this table, we additionally include the average (open loop) planning time per method and the average rate of safety violations. Safety violations are defined to occur where there are collisions, the robot hits its joint limits or there are torque limit errors. The open loop planning time for neural methods such as ours or M$\pi$Nets involves simply measuring the total time taken for rolling out the policy and test time optimization (TTO). We find that sampling-based planners in general never collide when executed. If they produce a safety violation, it is only because they find a trajectory that is infeasible for the robot to execute on the hardware, due to joint or torque limit errors. Neural motion planning methods have much higher collision rates, though \ours has a significantly lower collision rate than M$\pi$Nets, which we attribute to test-time optimization pruning out bad trajectories. We also note that not all collisions are created equal: some are slight, lightly grazing the environment objects while still achieving the goal, while others can be catastrophic, colliding heavily into the environment. In general, we found that our method tends to produce trajectories that may have slight collisions, though most of these are pruned out by TTO. With regards to planning time, M$\pi$Nets is the fastest method, as our method expends additional compute rolling out 100x more trajectories and then selecting the best one using SDF-based collision checking. 

\begin{table}[h!]
    \resizebox{\linewidth}{!}{%
        \begin{tabular}{lcccccc}
        \toprule
         & Bins ($\uparrow$) & Shelf ($\uparrow$) & Articulated ($\uparrow$) & Avg. Success Rate ($\uparrow$) & Avg. Planning Time ($\downarrow$) & Avg. Safety Viol. Rate ($\downarrow$)\\
        
        \multicolumn{2}{l}{\textit{Sampling-based Planning}:}\vspace{0.4em}\\
        \texttt{AIT*-80s~\cite{strub2020adaptively}} & 93.75 & 75 & 50.0 & 72.92 & 80 & \textbf{0}\\
        \texttt{AIT*-10s (fast)~\cite{strub2020adaptively}} & 75.0 & 37.5 & 25.0 & 45.83 & 10 & 2.1 \\
        \midrule
        
        \multicolumn{2}{l}{\textit{Neural}:}\vspace{0.4em}\\
        \texttt{M$\pi$Nets~\cite{fishman2023motion}} & 18.75 & 25.0 & 6.25 & 16.67 & \textbf{1.0} & 18.75  \\
        \midrule
        \textbf{\texttt{Ours}} & \textbf{100} & \textbf{100} & \textbf{87.5} & \textbf{95.83} & 3.9 & 4.2  \\ 
        \bottomrule
        \end{tabular}
        }
        \caption{\small 
        \ours performs best across tasks for free-hand motion planning, demonstrating greater improvement as the task complexity grows.}
        \label{tab:main detailed results}
\end{table}

\subsection{Detailed In-hand Motion Planning Results}
In this section, we extend the in-hand results shown in the main paper with additional baselines (AIT*-80s, AIT*-10s and M$\pi$Nets). For this evaluation (see Tab.~\ref{tab:detailed in hand results}, we consider two of the four in-hand motion planning objects, namely joystick and book. We find sampling-based methods are able to perform in-hand motion planning quite well, matching the performance of our base policy as well as our method without Objaverse data. We also see that M$\pi$Nets is unable to perform in-hand motion planning on any of the evaluated tasks. This is likely because that network was not trained on data with objects in-hand, demonstrating the importance of including in-hand data when training neural motion planners. Finally, there is a significant gap in performance between our method with and without test-time optimization; pruning out colliding trajectories at test time is crucial for achieving high success rates on motion planning tasks. 

\begin{table}[h!]
    \resizebox{\linewidth}{!}{%
        \begin{tabular}{lccccc}
        \toprule
        & Book ($\uparrow$) & Joystick ($\uparrow$) & Avg. Success Rate ($\uparrow$) & Avg. Planning Time ($\downarrow$) & Avg. Safety Viol. Rate ($\downarrow$)\\
        \multicolumn{2}{l}{\textit{Sampling-based Planning}:}\vspace{0.4em}\\
        \texttt{AIT*-80s~\cite{strub2020adaptively}} & 50 & 50		& 50	& 80	& \textbf{0}\\
        \texttt{AIT*-10s (fast)~\cite{strub2020adaptively}} 	& 25 & 50	& 37.5& 	10& 	\textbf{0} \\
        \midrule
        
        \multicolumn{2}{l}{\textit{Neural}:}\vspace{0.4em}\\
        \texttt{M$\pi$Nets~\cite{fishman2023motion}} & 0& 	0& 	0	& 1	& 37.5 \\
        \midrule
        \multicolumn{2}{l}{\textit{Ours}:}\vspace{0.4em} \\
        \texttt{Ours} (no TTO) 	& 25 & \textbf{75}	& 50	& \textbf{0.9} & 	50  \\ 
        \texttt{Ours} (no Objaverse) & 50 & 	50	& 50	& 3.9	& 50  \\ 
        \textbf{\texttt{Ours}} 	& \textbf{100} & \textbf{75}	& \textbf{87.5}	& 3.9 & 	12.5  \\ 
        \bottomrule
        \end{tabular}
        }
        \caption{\small 
        \ours performs best across tasks for in-hand motion planning, demonstrating greater improvement as the in-hand object becomes more challenging.}
        \label{tab:detailed in hand results}
\end{table}

\subsection{Test-time Optimization Analysis}

\begin{figure}[h!]
\centering
\includegraphics[width=\linewidth]{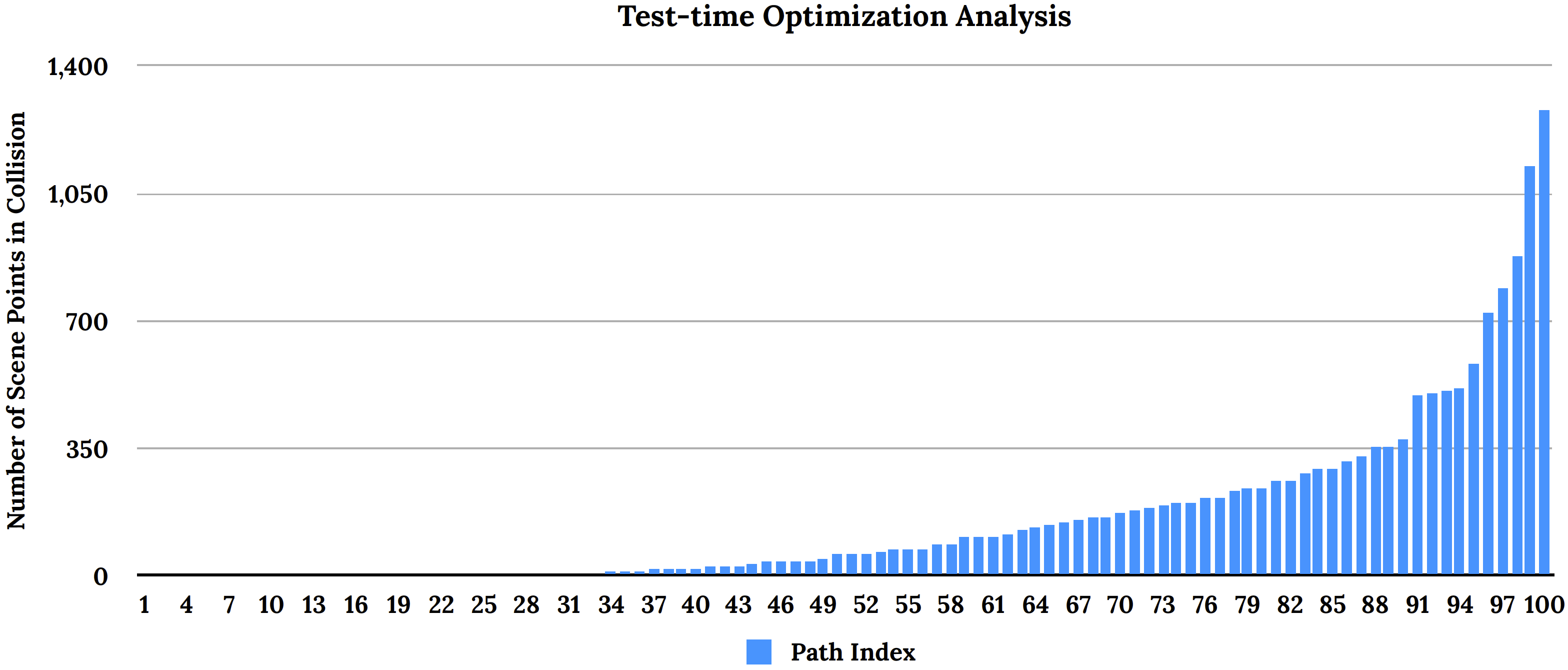}
\caption{\small \textbf{Test-time Optimization Analysis} For the Bins Scene 1 task, we plot the number of points in collision across 100 sampled trajectories from the model. 25\% of the trajectories are completely collision free and we select a trajectory execute from that subset.}
\label{fig:tto_analysis}
\end{figure}

To analyze what the test-time optimization procedure is doing, we first note that the base policy can sometimes produce slight collisions with the environment due to the imprecision of regression. As a result, when sampling from the policy, it is often likely that the policy will lightly graze objects which will count as failures when motion planning. We visualize a set of trajectories sampled from the policy here on our website for the real-world bins task. Observe that for some of the trajectories, the policy slightly intersects with the bin which would cause it to fail when executing in the real world, while for others it simply passes over the bin completely without colliding. We estimate the robot-scene intersection of all of these trajectories by comparing the robot SDF to the scene point-cloud and plot the range of values in Fig.~\ref{fig:tto_analysis}. We observe that 25\% of trajectories do not collide with the environment, and we select for those. In principle, one could further optimize by selecting the trajectory that is furthest from the scene (using the SDF). In practice, we did not find this necessary and that selecting the first trajectory among those with the fewest expected collisions performed quite well in our experiments.

\clearpage
\section{Ablations}
\label{app:ablations}

\begin{figure}[h!]
    \centering
    \includegraphics[height=50pt,width=\linewidth]{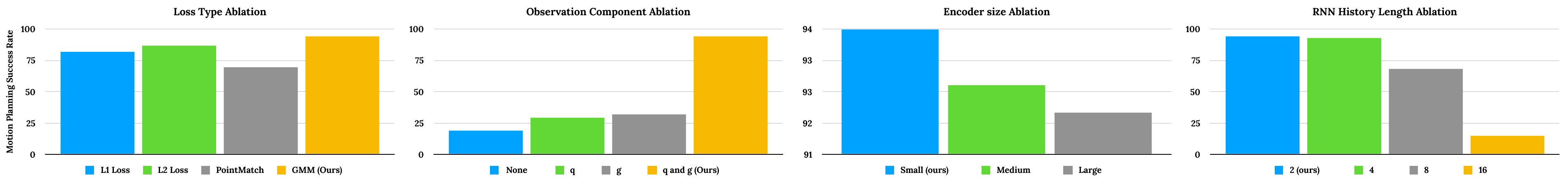}
    \caption{\small \textbf{Ablation Results} We evaluate four different components of \ours, loss type (\textit{left}), observation components (\textit{middle left}), encoder sizes (\textit{middle right}), and RNN history length (\textit{right}). We validate that our design decisions produce measurable improvements in motion planning success rates. }
    \label{fig:ablations}
\end{figure}
We run additional ablations analyzing components of our method in simulation using a subset of our dataset (100K trajectories) and include additional details for experiments discussed in the main paper.

\textbf{Loss Types}
For training objective, we evaluate 4 different options: GMM log likelihood (ours), MSE loss, L1 loss, and PointMatch loss (M$\pi$Nets). PointMatch loss involves computing the l2 distance between the goal and the predicted end-effector pose using 1024 key-points. We plot the results on held out scenes in Fig.~\ref{fig:ablations}. We find that GMM (ours) outperforms L2 loss, L1 loss, and PointMatch Loss (M$\pi$Nets) by (7\%, 12\%, and 24\%) respectively. One reason this may be the case is that sampling-based motion planners produce highly multi-modal trajectories: they can output entirely different trajectories for the same start and goal pair when sampled multiple times. Since Gaussian Mixture Models are generally more capable of capturing multi-modal distributions, they can hence fit our dataset well. At the same time, the PointMatch~\cite{fishman2023motion} loss struggles significantly on our data: it cannot distinguish between 0 and 180 degree flipped end-effector orientations, resulting in many failures due to incorrect end-effector orientations.

\textbf{Observation Components}
We evaluate whether our choice of observation components impacts the \ours's performance. In theory, the network should be able to learn as well from the point-cloud alone as when the proprioception is included, as the point-cloud contains a densely sampled point-cloud of the current and goal robot configurations. However, in practice, we find that this is not the case. Instead, removing either $q$ or $g$ or both severely harms performance as seen in Fig.~\ref{fig:ablations}. We hypothesize that including the proprioception provides a richer signal for the correct delta action to take.

\textbf{RNN History Length}
In our experiments, we chose a history length of 2 for the RNN, after sweeping over values of 2, 4, 8, 16 based on performance. From Fig.~\ref{fig:ablations} we see history length 2 achieves the best performance at 94\%, while using lengths 4, 8 and 16 achieve progressively decreasing success rates (92.67, 68, 14.67). One possible reason for this is that since point-clouds are already very dense representations that cover the scene quite well, the partial observability during training time is fairly low. A shorter history length also leads to faster training, due to smaller batches and fewer RNN unrolling steps.

\textbf{Encoder Size}
Finally, we briefly evaluate whether encoder size is important when training large-scale neural motion planners. We train 3 different size models: small (4M params), medium (8M params) and large (16M params). From the results in Fig.~\ref{fig:ablations}, we find that the encoder size does not affect performance by a significant margin (94\%, 93\%, 92\%) respectively and that the smallest model in fact performs best. Based on these results, we opt to use the small, 4M param model in our experiments.

\begin{table}[h]
\centering
\scriptsize
\resizebox{\linewidth}{!}{%
\begin{tabular}{@{}cccc@{}}
\toprule
\textbf{\ours-MLP} & \textbf{\ours-LSTM} & \textbf{\ours-Transformer} & \textbf{\ours-ACT} \\
\midrule 
65.0 & 82.5 & \textbf{85.0} & 47.5 \\
\bottomrule
\end{tabular}
}
\vspace{5pt}
\caption{\small Ablation of different architecture choices for the action decoder. We find that LSTMs and Transformers comparably while LSTMs boast faster inference times.}
\label{tab:arch abl}
\vspace{-10pt}
\end{table}

\textbf{Architecture Ablation}
In this experiment, we evaluate how different sequence modelling methods (Transformers and ACT~\cite{zhao2023learning}) and simpler action decoders such as MLPs compare against our design choice of using an LSTM. All methods are trained with the same dataset (of 1M trajectories), with the same encoder and GMM output distribution (with the exception of ACT which uses an L1 loss as per the ACT paper). We then evaluate them on held out motion planning tasks (Fig.~\ref{tab:arch abl} which are replicas of our real-world tasks (Bins and Shelf). We note several findings: 1) ACT performs poorly, largely due to its design choice of using an L1 loss which prevents it from handling planner multi-modality effectively, 2) \ours with an MLP action decoder also performs significantly worse than LSTMs and Transformers, as it is unable to use history information effectively to reason about the next action 3) Transformers and LSTMs perform similarly, with the Transformer variant performing marginally better, but with significantly slower inference time (2x). Hence we opt to use LSTM policies for our experimental evaluation, but certainly our method is amenable to any choice of sequence modeling architecture that performs well and has fast inference.

\begin{table}[h]
\centering
\scriptsize
\begin{tabular}{@{}ccc@{}}
\toprule
\textbf{\ours-MotionBenchMaker} & \textbf{\ours-M$\pi$Nets} & \textbf{\ours} \\
\midrule 
0 & 32.5 & 82.5 \\
\bottomrule
\end{tabular}
\vspace{5pt}
\caption{\small Comparing different methods for generating datasets for motion planning. We find that policies trained on our data generalize best to held out scenes.}
\label{tab:datagen abl}
\vspace{-10pt}
\end{table}
\textbf{Dataset Ablation}
Finally, we evaluate the quality of different dataset generation approaches for producing generalist neural motion planners. We do so by training policies on three different datasets (\ours, M$\pi$Nets~\cite{fishman2023motion}, and MotionBenchMaker~\cite{chamzas2021motionbenchmaker}) and evaluated on held out motion planning tasks in simulation. We train each model to convergence for 10K epochs and then execute trajectories on two held out tasks that mirror our real world tasks: RealBins and RealShelf. For fairness, we do not include any Objaverse meshes in these tasks, since MPiNets and MotionBenchMaker only have primitive objects. Still, we find that our dataset performs best by a wide margin (Tab.~\ref{tab:datagen abl}).
In general, we found that policies trained on MotionBenchMaker do not generalize well. As mentioned in the related works section, this dataset lacks the realism and diversity necessary to train policies that can generalize to held out motion planning scenes. 

\section{Procedural Scene Generation Details}
\label{app:procgen}

In this section we provide additional details regarding the data generation methods we develop for training large scale neural motion planners. 

\subsection{Procedural Scene Generation}

We formalize our procedural scene generation as a composition of randomly generated parameteric assets and sampled Objaverse meshes in Alg.~\ref{alg:scene_gen}

\textbf{Objaverse sampling details} 
The Objaverse are sampled in the task-relevant sampling location of the programmatic asset(s) in the scene, such as between shelf rungs, inside cubbies or within cabinets. Similar to the programmatic assets, these Objaverse assets are also sampled from a category generator $X_{obj}(\textbf{p})$. Here the parameter $p$ specifies the size, position, orientation of the object as well as task-relevant sampling location of the object in the scene, such as between shelf rungs, inside cubbies or within cabinets. 
As discussed in the main paper, we propose an approach that iteratively adds assets to a scene by adjusting their position using the effective collision normal vector, computed from the existing assets in the scene. We detail the steps for doing this in Alg.~\ref{alg:scene_gen}.

\subsection{Motion Planner Experts}
We use three techniques to improve the data generation throughput when imitating motion planners at scale.

\textbf{Hindsight Relabeling}
Tight-space to tight-space problems are the most challenging, particularly for sampling-based planners, often requiring significant planning time (up to 120 seconds) for the planner to find a solution. 
For some problems, the expert planner is unable to find an exact solution and instead produces approximate solutions. Instead of discarding these, note that we use a goal-conditioned imitation learning framework, where we can simply execute the trajectories in simulation and relabel the observed final state as the new goal.

\textbf{Reversibility}
We further improve our data generation throughput by observing that since motion planners inherently produce collision-free paths, the process is reversible, at least in simulation. This allows us to double our data throughput by reversing expert trajectories and re-calculating delta actions accordingly. Additionally, for a neural motion planner to be useful for practical manipulation tasks, it must be able to generate collision free plans for the robot even when it is holding objects. 
To enable such functionality, we augment our data generation process with trajectories 
where objects are spawned between the grippers of the robot end effector. There are transformed along with the end-effector during planning in simulation. We consider the object as part of the robot for collision checking and for the sake of our visual observations. 
In order to handle diverse objects that the robot might have to move with at inference time, we perform significant randomization of the in-hand object that we spawn in simulation. Specifically, we sample this object from the primitive categories of boxes, cylinders or spheres, or even from Objaverse meshes of everyday articles. We randomize the scale of the object between 3 and 30 cm along the longest dimension, and sample random starting locations within a 5cm cube around the end-effector mid-point between grippers.

\textbf{Smoothing}
Importantly, we found that naively imitating the output of the planner performs poorly in practice as the planner output is not well suited for learning. Specifically, plans produced by AIT* often result in way-points that are far apart, creating large action jumps and sparse data coverage, making it difficult to for networks to fit the data. To address this issue, we perform smoothing using cubic spline interpolation while enforcing velocity and acceleration limits. The implementation from M$\pi$Nets performs well in practice, smoothing to a fixed 50 timesteps with a max spacing of 0.1 radians. In general, we found that smoothing is crucial for learning performance as it ensures the maximum action size is small and thus easier for the network to fit to.

\subsection{Data Pipeline Parameters and Compute}
In Table~\ref{tab:datagen params}, we provide a detailed list of all the parameters used in generating the data to train our model. 

\textbf{Compute}
In order to collect a vast data of motion planning trajectories, we parallelize data collection across a cluster of 2K CPUs. It takes approximately 3.5 days to collect 1M trajectories. 

\section{Network Training Details}
\label{app:network details}

We first describe additional details regarding our neural policy, and then discuss how it is trained. 
Following the design decisions of M$\pi$Nets~\cite{fishman2023motion}, we construct a segmented point-cloud for the robot, consisting of the robot point-cloud, the target goal robot point-cloud and the obstacle point-cloud. Here we note two key differences from M$\pi$Nets: 1) our network conditioned on the target joint angles, while M$\pi$Nets only does so through the segmented point-cloud, 2) we condition on the target joint angles, not end-effector pose, decisions that we found improved adherence to the overall target configuration. For in-hand motion planning, we extend this representation by considering the object in-hand as part of the robot for the purpose of segmentation. 

We include a hyper-parameter list for our neural motion planner in Table~\ref{tab:model hyper params}. We train a 20M parameter neural network across our dataset of 1M trajectories. The PointNet++ encoder is 4M parameters and outputs an embedding of dimension 1024. We concatenate this embedding with the encoded $q_t$ and $g$ vectors and pass this into the 16M parameter LSTM decoder. The decoder outputs weights, means, and standard deviations of the 5 GMM modes. We then train the model with negative log likelihood loss for 4.5M gradient steps, which takes 2 days on a 4090 GPU with batch size of 16.

\clearpage
\onecolumn
\begin{longtable}{l|l}

\centering

\textbf{Hyper-parameter} & \textbf{Value} \\ \hline
\endfirsthead

\textbf{Hyper-parameter} & \textbf{Value} \\ \hline
\endhead
\multicolumn{2}{c}{\textbf{General Motion Planning Parameters}} \\ \hline
collision checking distance & 1cm \\
\rowcolor{Gray}
tight space configuration ratio & 50\% \\ 
dataset size & 1M trajectories \\ 
\rowcolor{Gray}
minimum motion planning time & 20s \\ 
maximum motion planning time & 80s \\
\hline
\multicolumn{2}{c}{\textbf{General Obstacle Parameters}} \\ \hline
in hand object ratio & 0.5 \\
\rowcolor{Gray}
in hand object size range & [[0.03, 0.03, 0.03], [0.3, 0.3, 0.3]] \\ 
in hand object xyz range & [[-0.05, -0.05, 0.], [0.05, 0.05, 0.05]] \\ 
\rowcolor{Gray}
min obstacle size & 0.1 \\ 
max obstacle size & 0.3 \\ 
\rowcolor{Gray}
table dim ranges & [[0.6, 1], [1.0, 1.5], [0.05, 0.15]] \\ 
table height range & [-0.3, 0.3] \\ 
\rowcolor{Gray}
num shelves range & [0, 3] \\ 
num open boxes range & [0, 3] \\ 
\rowcolor{Gray}
num cubbys range & [0, 1] \\ 
num microwaves range & [0, 3] \\ 
\rowcolor{Gray}
num dishwashers range & [0, 3] \\ 
num cabinets range & [0, 3] \\ 
\hline
\multicolumn{2}{c}{\textbf{Objaverse Mesh Parameters}} \\ \hline
scale range & [0.2, 0.4] \\ 
\rowcolor{Gray}
x pos range & [0.2, 0.4] \\ 
y pos range & [-0.4, 0.4] \\ 
\rowcolor{Gray}
number of mesh objects per programmatic asset & [0, 3] \\ 
number of mesh objects on the table & [0, 5] \\ \hline
\multicolumn{2}{c}{\textbf{Table Parameters}} \\ \hline
width range & [0.8, 1.2] \\ 
\rowcolor{Gray}
depth range & [0.4, 0.6] \\ 
height range & [0.35, 0.5] \\ 
\rowcolor{Gray}
thickness range & [0.03, 0.07] \\ 
leg thickness range & [0.03, 0.07] \\ 
\rowcolor{Gray}
leg margin range & [0.05, 0.15] \\ 
position range & [[0, 0.8], [-0.6, 0.6]] \\ 
\rowcolor{Gray}
z axis rotation range & [0, 3.14] \\ \hline
\multicolumn{2}{c}{\textbf{Shelf Parameters}} \\ \hline
width range & [0.5, 1] \\ 
\rowcolor{Gray}
depth range & [0.2, 0.5] \\ 
height range & [0.5, 1.2] \\ 
\rowcolor{Gray}
num boards range & [3, 5] \\ 
board thickness range & [0.02, 0.05] \\ 
\rowcolor{Gray}
backboard thickness range & [0.0, 0.05] \\ 
num vertical boards range & [0, 3] \\ 
\rowcolor{Gray}
num side columns range & [0, 4] \\ 
column thickness range & [0.02, 0.05] \\ 
\rowcolor{Gray}
position range & [[0, 0.8], [-0.6, 0.6]] \\ 
z axis rotation range & [-1.57, 0] \\ \hline
\multicolumn{2}{c}{\textbf{Open Box Parameters}} \\ \hline
width range & [0.2, 0.7] \\ 
\rowcolor{Gray}
depth range & [0.2, 0.7] \\ 
height range & [0.3, 0.5] \\
\rowcolor{Gray}
thickness range & [0.02, 0.06] \\ 
front scale range & [0.6, 1] \\ 
\rowcolor{Gray}
position range & [[0.0, 0.8], [-0.6, 0.6]] \\ 
z axis rotation range & [-1.57, 0.0] \\ \hline
\multicolumn{2}{c}{\textbf{Cubby Parameters}} \\ \hline
cubby left range & [0.4, 0.1] \\ 
\rowcolor{Gray}
cubby right range & [-0.4, 0.1] \\ 
cubby top range & [0.85, 0.35] \\ 
\rowcolor{Gray}
cubby bottom range & [0.0, 0.1] \\ 
cubby front range & [0.8, 0.1] \\ 
\rowcolor{Gray}
cubby width range & [0.35, 0.2] \\ 
cubby horizontal middle board z axis shift range & [0.45, 0.1] \\ 
\rowcolor{Gray}
cubby vertical middle board y axis shift range & [0.0, 0.1] \\ 
board thickness range & [0.02, 0.01] \\ 
\rowcolor{Gray}
external rotation range & [0, 1.57] \\ 
internal rotation range & [0, 0.5] \\ 
\rowcolor{Gray}
num shelves range & [3, 5] \\ \hline
\multicolumn{2}{c}{\textbf{Microwave Parameters}} \\ \hline
width range & [0.3, 0.6] \\ 
\rowcolor{Gray}
depth range & [0.3, 0.6] \\ 
height range & [0.3, 0.6] \\ 
\rowcolor{Gray}
thickness range & [0.01, 0.02] \\ 
display panel width range & [0.05, 0.15] \\ 
\rowcolor{Gray}
distance range & [0.5, 0.8] \\ 
external z axis rotation range & [-2.36, -0.79] \\ 
\rowcolor{Gray}
internal z axis rotation range & [-0.15, 0.15] \\ \hline
\multicolumn{2}{c}{\textbf{Dishwasher Parameters}} \\ \hline
width range & [0.4, 0.6] \\ 
\rowcolor{Gray}
depth range & [0.3, 0.4] \\ 
height range & [0.5, 0.7] \\ 
\rowcolor{Gray}
control panel height range & [0.1, 0.2] \\ 
foot panel height range & [0.1, 0.2] \\ 
\rowcolor{Gray}
wall thickness range & [0.01, 0.02] \\ 
opening angle range & [0.5, 1.57] \\ 
\rowcolor{Gray}
distance range & [0.6, 1.0] \\ 
external z axis rotation range & [-2.36, -0.79] \\ 
\rowcolor{Gray}
internal z axis rotation range & [-0.15, 0.15] \\ \hline
\multicolumn{2}{c}{\textbf{Cabinet Parameters}} \\ \hline
width range & [0.5, 0.8] \\ 
\rowcolor{Gray}
depth range & [0.25, 0.4] \\ 
height range & [0.6, 1.0] \\ 
\rowcolor{Gray}
wall thickness range & [0.01, 0.02] \\ 
left opening angle range & [0.7, 1.57] \\ 
\rowcolor{Gray}
right opening angle range & [0.7, 1.57] \\ 
distance range & [0.6, 1.0] \\ 
\rowcolor{Gray}
external z axis rotation range & [-2.36, -0.79] \\ 
internal z axis rotation range & [-0.15, 0.15] \\ 

\caption{\small \textbf{Data Generation Hyper-parameters} We provide a detailed list of hyper-parameters used to procedurally generate a vast variety of scenes in simulation. }
\label{tab:datagen params}
\end{longtable}
\clearpage
\twocolumn

\clearpage

\begin{table*}[h!]
\centering
\begin{tabular}{l|l}
\hline
\textbf{Hyper-parameter} & \textbf{Value} \\ \hline
PointNet++ Architecture & 
\begin{minipage}[t]{0.5\textwidth}
\begin{verbatim}
PointnetSAModule(
    npoint=128,
    radius=0.05,
    nsample=64,
    mlp=[1, 64, 64, 64],
)
PointnetSAModule(
    npoint=64,
    radius=0.3,
    nsample=64,
    mlp=[64, 128, 128, 256],
)
PointnetSAModule(
    nsample=64,
    mlp=[256, 512, 512],
)
MLP(
    Linear(512, 2048),
    GroupNorm(16, 2048),
    LeakyReLU,
    Linear(2048, 1024),
    GroupNorm(16, 1024),
    LeakyReLU,
    Linear(1024, 1024)
)
\end{verbatim}
\end{minipage} \\ 
\rowcolor{Gray}
LSTM & 1024 hidden dim, 2 layers \\ 
Inputs & $q_t$, $g$, $PCD_t$ \\ 
\rowcolor{Gray}
Batch Size & 16 \\ 
Learning Rate & $0.0001$ \\ 
\rowcolor{Gray}
GMM & 5 modes \\
Sequence Length (seq length) & 2 \\ \hline
\multicolumn{2}{c}{\textbf{Point Cloud Parameters}} \\ \hline
Number of Robot / Goal Point-cloud Points & 2048 \\ 
\rowcolor{Gray}
Number of Obstacle Point-cloud Points & 4096 \\ 
\end{tabular}
\caption{Hyper-parameters for the model}
\label{tab:model hyper params}
\end{table*}

\clearpage
\section{Real World Setup Details}
\label{app:real world}
In this section, we describe our real world robot setup and tasks in detail and perform analysis on the perception used for operating our policies. 

\subsection{Real Robot Setup}
\begin{figure}
\begin{algorithm}[H]
\caption{Open-Loop Execution of \ours}
\label{alg:open loop execution}
\begin{algorithmic}[1]
\State \textbf{Input:} \ours $\pi_{\theta}$, segmentor $\mathcal{S}$, initial angles $q_0$, scene point-cloud $PCD_{full}$, goal $g$, horizon $H$
\State \textbf{Output:} Executed trajectory on the robot
\State \textbf{Initialize: } Timestep $t \leftarrow 0$
\State \textbf{Initialize: } Trajectory $\tau \leftarrow \{\}$
\State $PCD_0 \leftarrow \mathcal{S}(PCD_{full}) \cup PCD_{q_0} \cup PCD_{g}$
\While{goal $g$ not reached $\And t < H$}
    \State $a_t \sim \pi_{\theta}(q_{t-1}, PCD_{t-1}, g)$
    \State $q_t \leftarrow q_{t-1} + a_t$
    \State $PCD_t \leftarrow (PCD_{t_1} \setminus PCD_{q_{t-1}}) \cup PCD_{q_t}$
    \State $t \leftarrow t + 1$
    \State $\tau \leftarrow \tau + a_t$
\EndWhile
\State Execute the $\tau$ open loop on the robot.
\end{algorithmic}
\end{algorithm}
\end{figure}

\textbf{Hardware} For all of our experiments, we use a Franka Emika Panda Robot, which is a 7 degree of freedom manipulator arm. We control the robot using the manimo library (\href{https://github.com/AGI-Labs/manimo}{https://github.com/AGI-Labs/manimo}) and perform all of experiments using their joint position controller with the default PD gains. The robot is mounted to a fixed base pedestal behind a desk of size .762m by 1.22m with variable height. For sensing, we use four extrinsically calibrated depth cameras, Intel Realsense 435 / 435i, placed around the scene in order to accurately capture the environment. We project the depth maps from each camera into 3D and combine the individual point-clouds into a single scene representation. 
We then post-process the point-cloud by cropping it to the workspace, filtering outliers and denoising, and sub-sampling a set of 4096 points. This processed point-cloud is then used as input to the policy.

\begin{figure}[t]
    \centering
    \vspace{-.2in}
    \includegraphics[width=.45\linewidth]{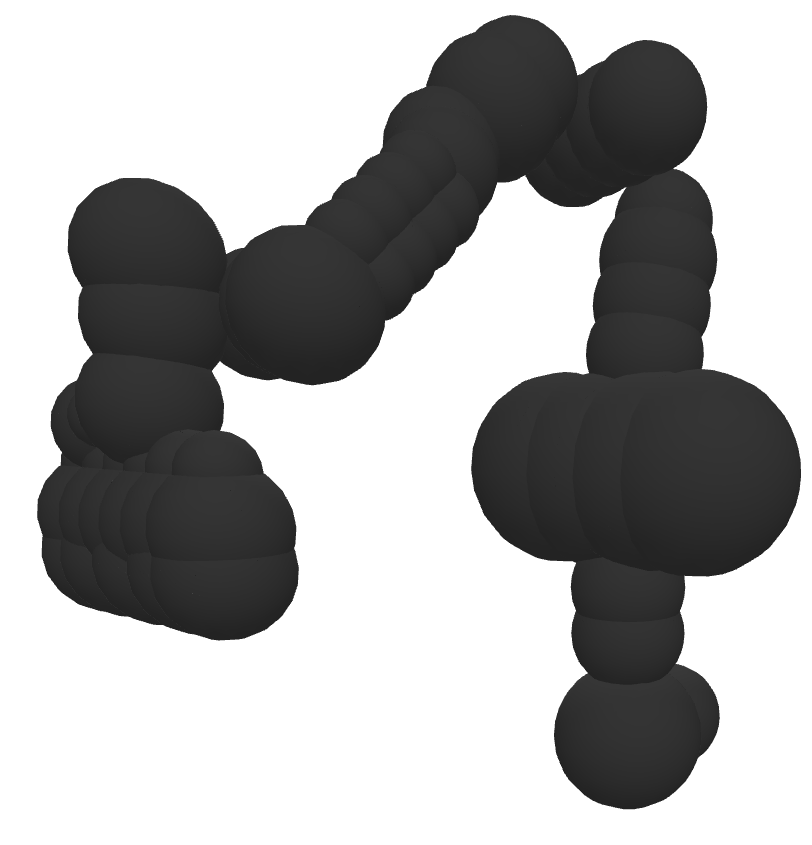}
    \includegraphics[width=.45\linewidth]{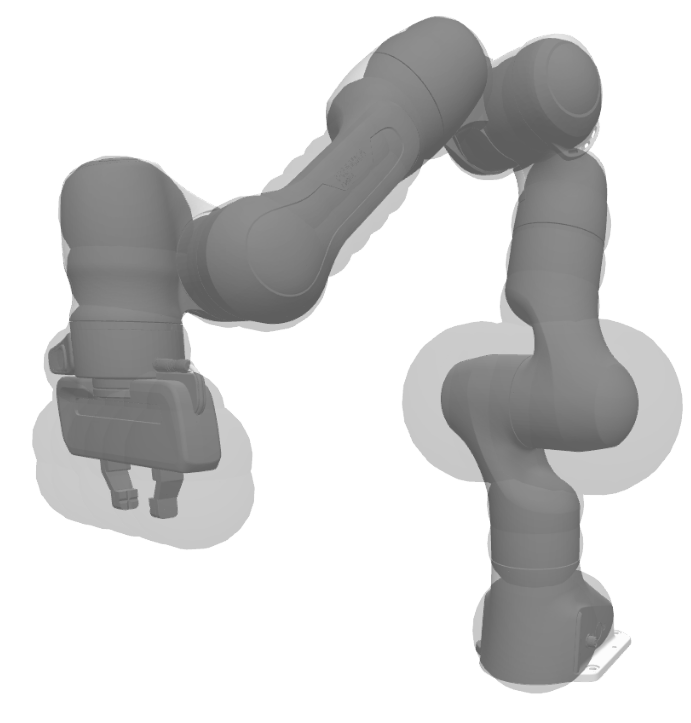}
    \caption{\small We visualize the spherical representation on the left and overlay it on the robot mesh on the right.}
    \label{fig:spherical repr vis}
    \vspace{-.1in}
\end{figure}

\textbf{Representation Collision Checking and Segmentation} In order to perform real world collision checking and robot point-cloud segmentation, we require a representation of the robot to check intersections with the scene (collision checking) and to filter out robot points from the scene point-cloud (segmentation). While the robot mesh is the ideal candidate for these operations, it is far too slow to run in real time. Instead, we approximate the robot mesh as spheres (visualized in Fig.~\ref{fig:spherical repr vis}) as we found this performs well in practice while operating an order of magnitude faster. We use 56 spheres in total to approximate the links of the robot as well as the end-effector and gripper. These have radii ranging from 2cm to 10cm and are defined relative to the center of mass of the link. This representation is a conservative one: it encapsulates the robot mesh, which is desirable for segmentation as this helps account for sensing errors which would place robot points outside of the robot mesh.

\textbf{Robot Segmentation} In order to perform robot segmentation in the real world, we use the spherical representation to filter out robot points in the scene, so only the obstacle point-cloud remains. Doing so requires computing the Signed Distance Function (SDF) of the robot representation and then checking the scene point-cloud against it, removing points from the point-cloud in which SDF value is less than threshold $\epsilon$. For the spherical representation, the SDF computation is efficient: for a sphere with center $C$ and radius $r$, the SDF of point x is simply $ ||x - C||_2 - r$. In our experiments, we use a threshold $\epsilon$ of 1cm. We then replace the removed points with points sampled from the robot mesh of the robot. This is done by pre-sampling a robot point-cloud from the robot mesh at the default configuration, then performing forward kinematics using the current joint angles $q_t$ and transforming the robot point-cloud accordingly. Replacing the real robot point-cloud with this sampled point-cloud ensures that the only difference between sim and real is the obstacle point-cloud. 

\textbf{Real-world Collision Checking} Given the SDF, collision checking is also straightforward, we denote the robot in collision if any point in the scene point-cloud (this is after robot segmentation) has SDF value less than 1cm. Note this means that first state is by definition collision free. Also, this technique will not hold if performing closed loop planning, in that case this method would always denote the state as collision free as the points with SDF value less than 1cm would be segmented out for each intermediate point-cloud.

\textbf{Open Loop Deployment} For open-loop execution of neural motion planners, we execute the following steps: 1) generate the segmented point-cloud at the first frame, 2) predict the next trajectory way-point by computing a forward pass through the network and sampling an action, 3) update the current robot point-cloud with mesh-sampled point-cloud at the predicted way-point, and 4) repeat until goal reaching success or maximum rollout length is reached. The entire trajectory is then executed on the robot after the rollout. Please see Alg.~\ref{alg:open loop execution} for a more detailed description of our open-loop deployment method.

\subsection{Tasks}

\begin{figure*}[ht!]
\centering
\begin{subfigure}[b]{0.24\linewidth}
    \includegraphics[width=\linewidth]{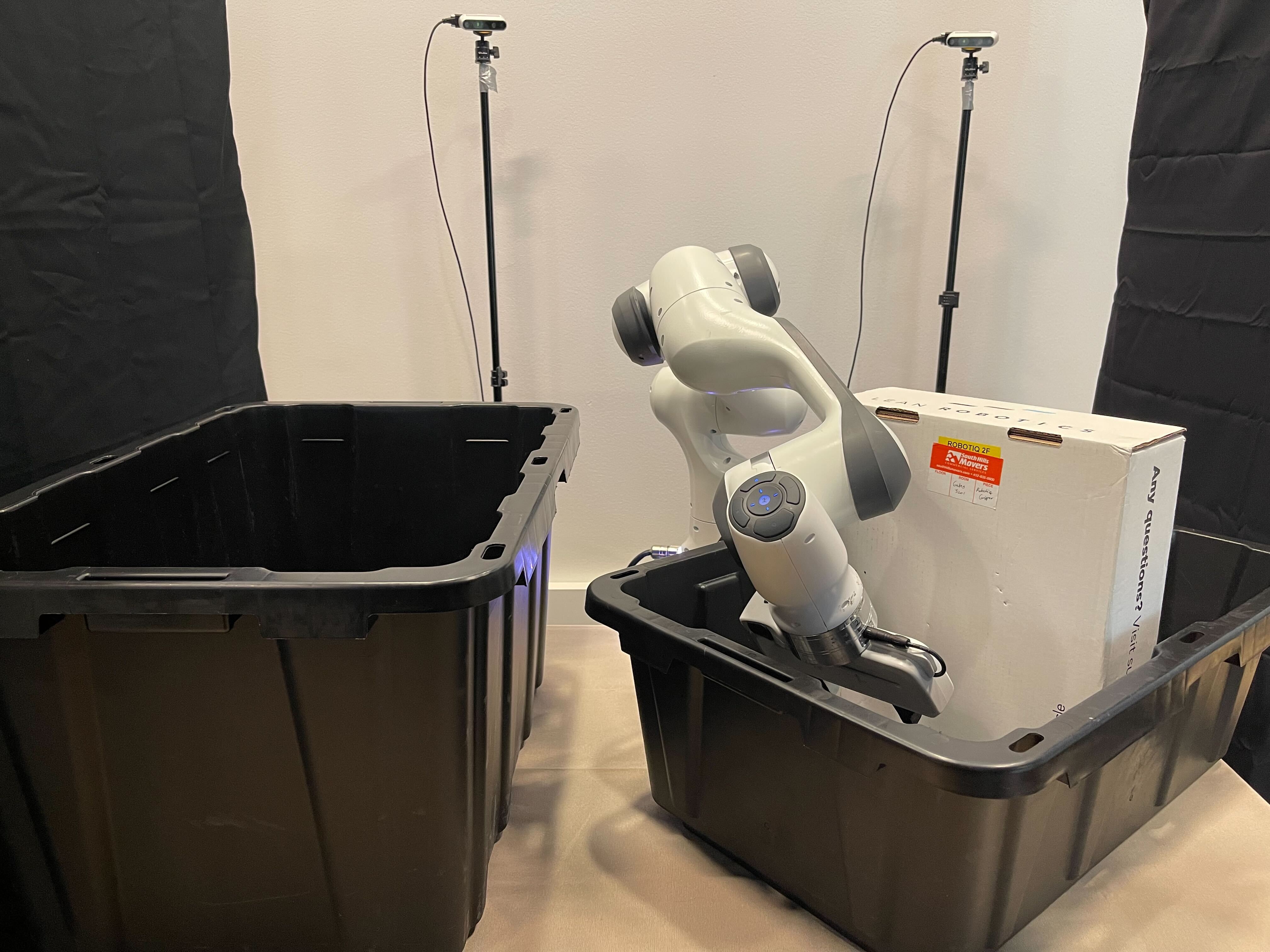}
    \caption{\small Bins Scene 1}
\end{subfigure}
\begin{subfigure}[b]{0.24\linewidth}
    \includegraphics[width=\linewidth]{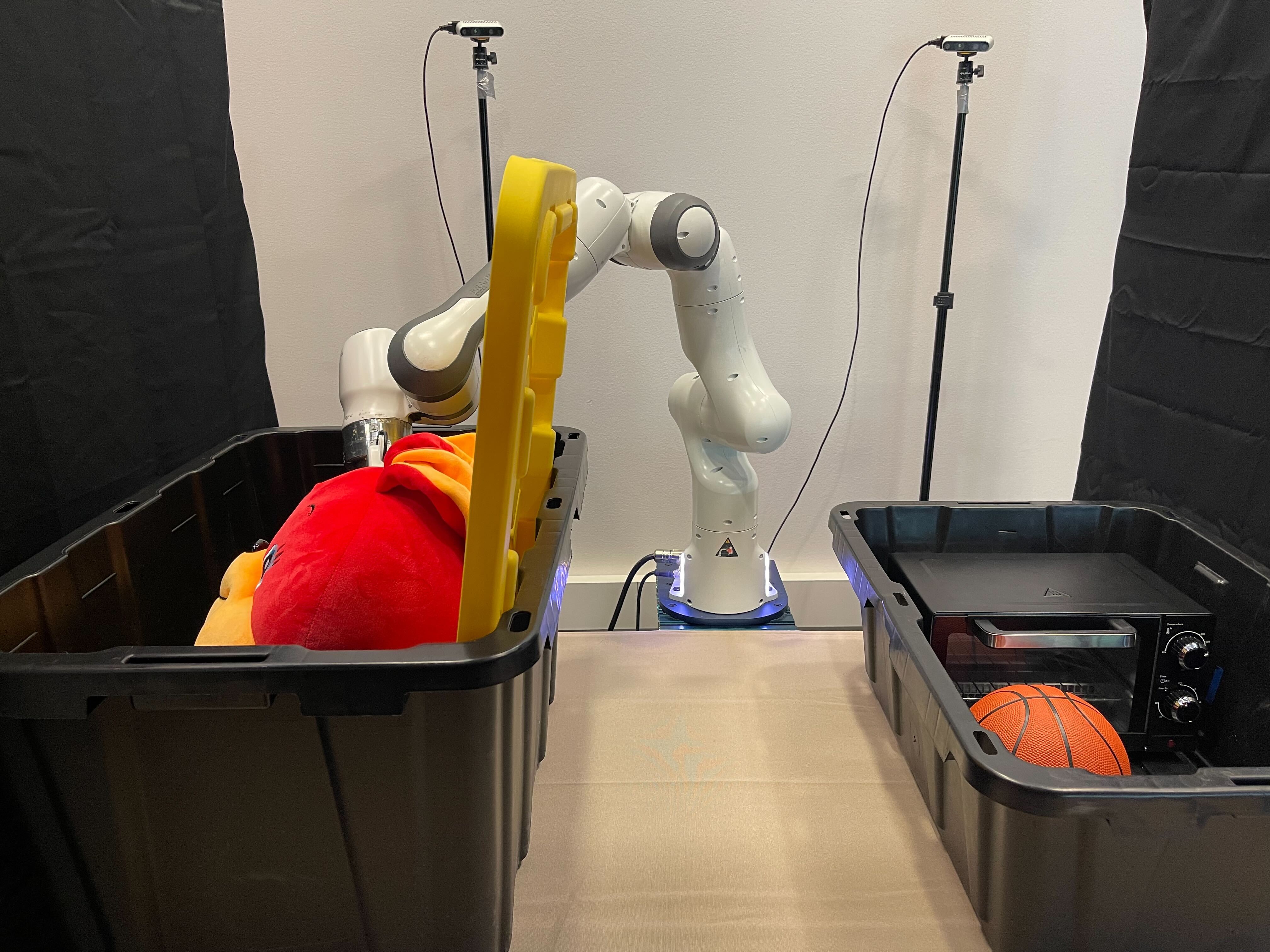}
    \caption{\small Bins Scene 2}
\end{subfigure}
\begin{subfigure}[b]{0.24\linewidth}
    \includegraphics[width=\linewidth]{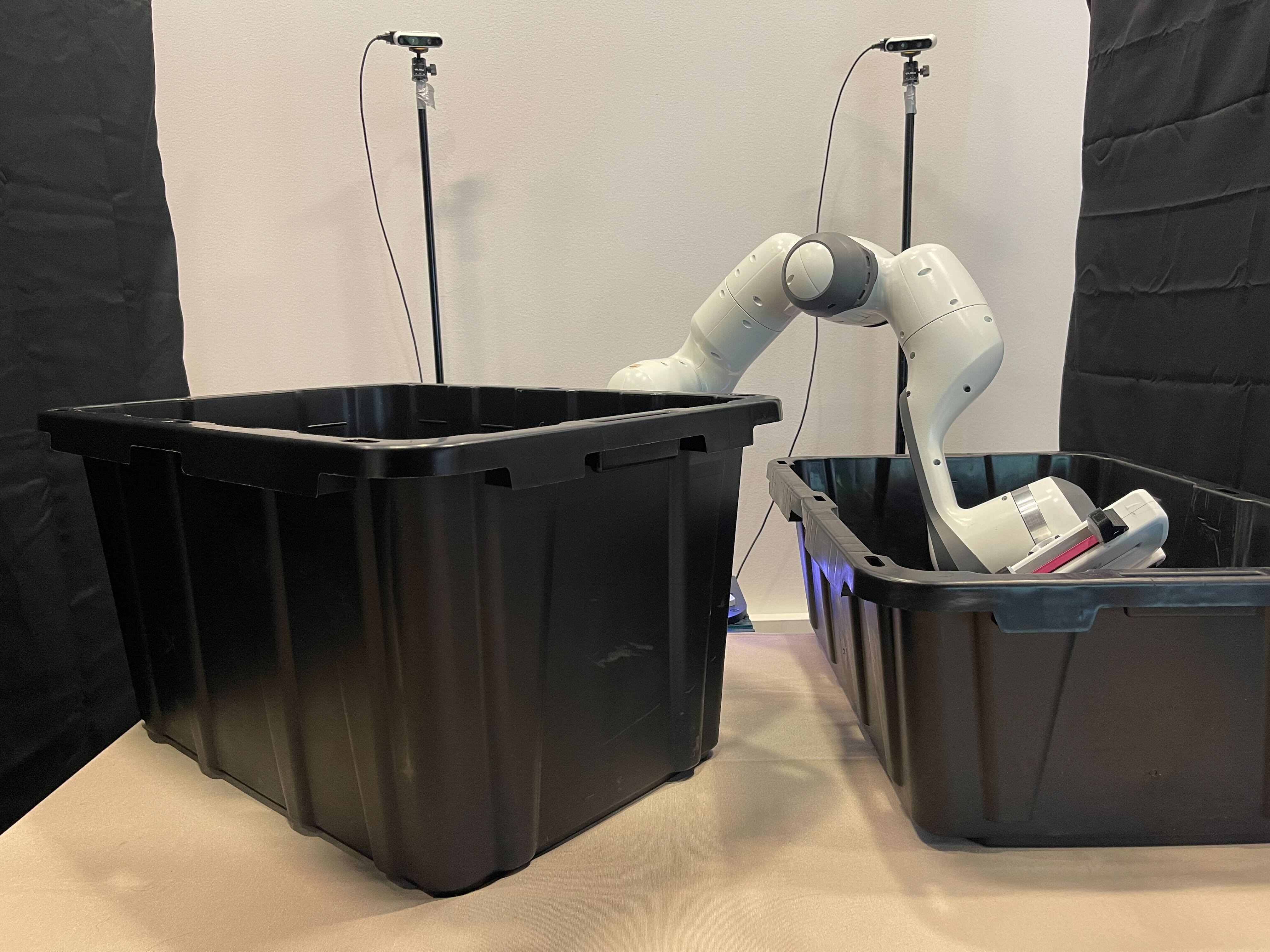}
    \caption{\small Bins Scene 3}
\end{subfigure}
\begin{subfigure}[b]{0.24\linewidth}
    \includegraphics[width=\linewidth]{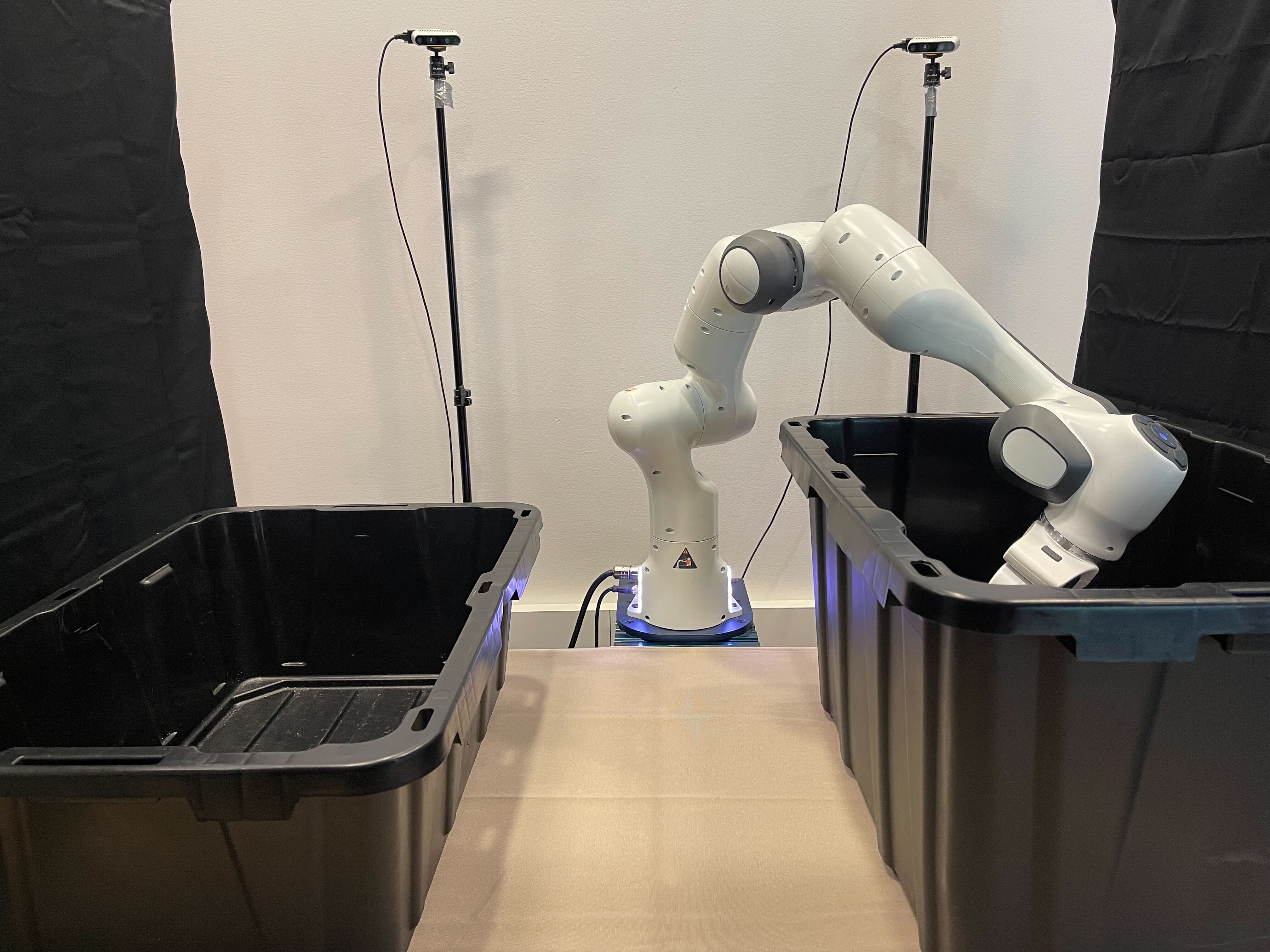}
    \caption{\small Bins Scene 4}
\end{subfigure}
\begin{subfigure}[b]{0.24\linewidth}
    \includegraphics[width=\linewidth]{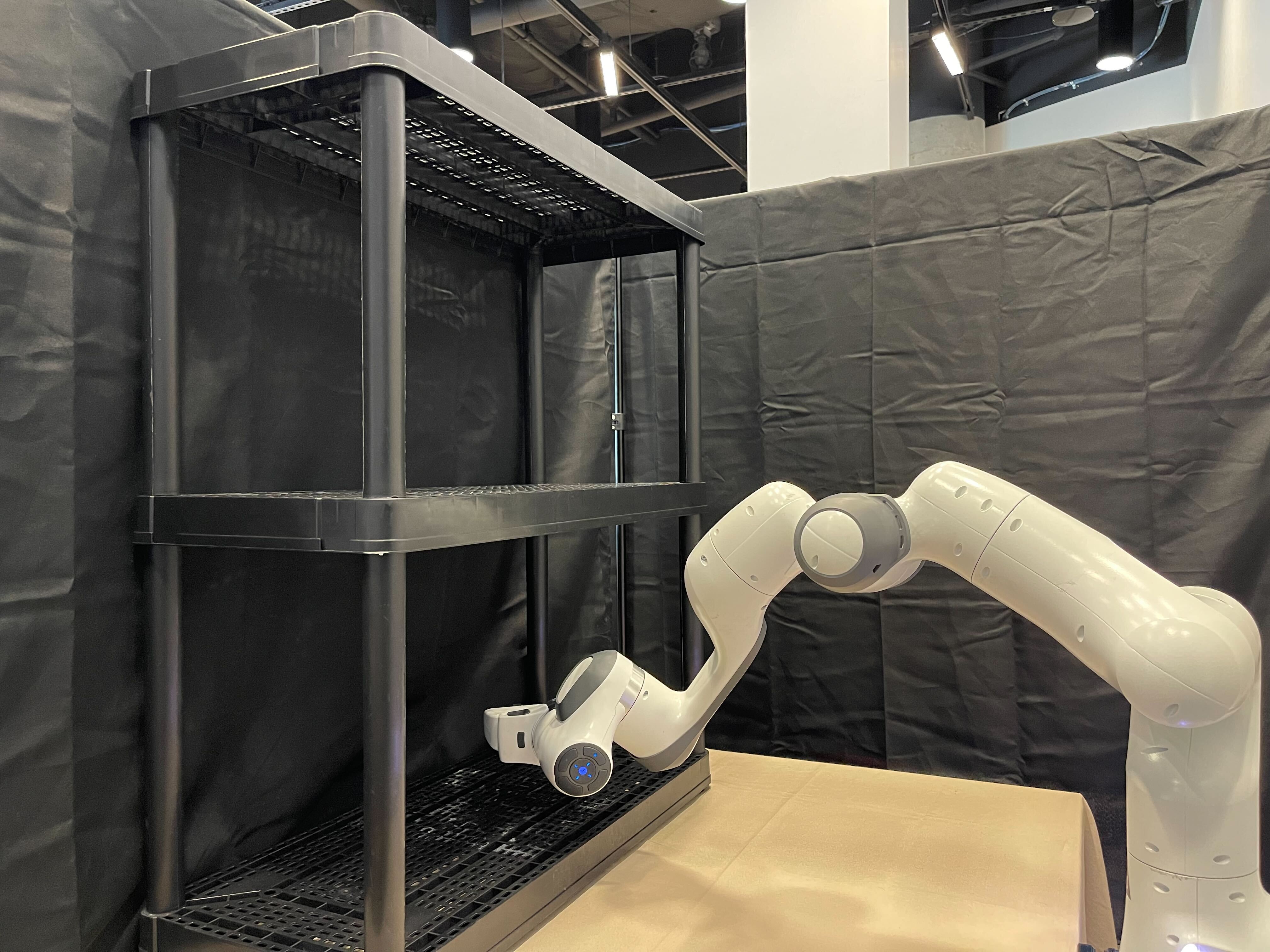}
    \caption{\small Shelf Scene 1}
\end{subfigure}
\begin{subfigure}[b]{0.24\linewidth}
    \includegraphics[width=\linewidth]{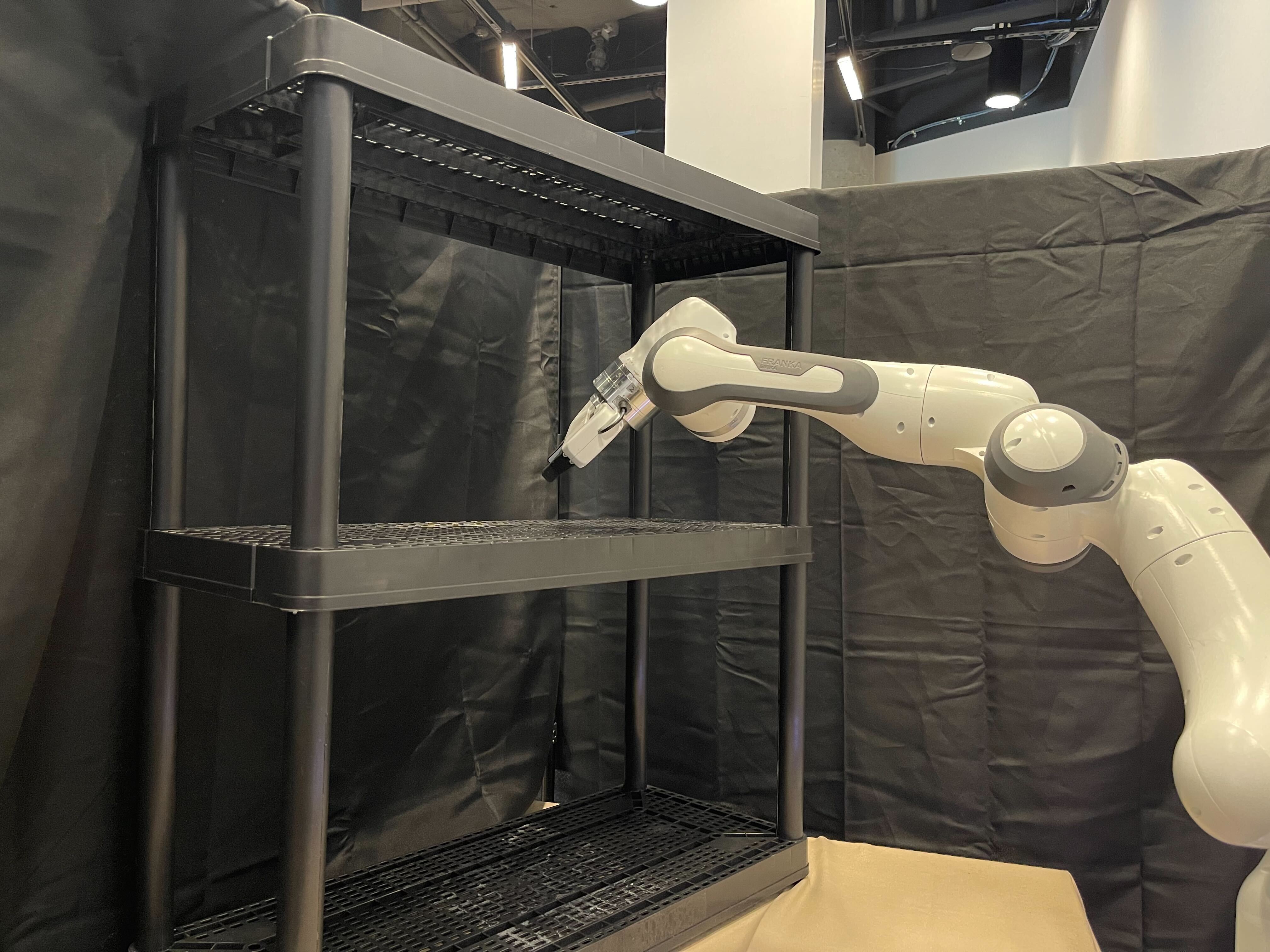}
    \caption{\small Shelf Scene 2}
\end{subfigure}
\begin{subfigure}[b]{0.24\linewidth}
    \includegraphics[width=\linewidth]{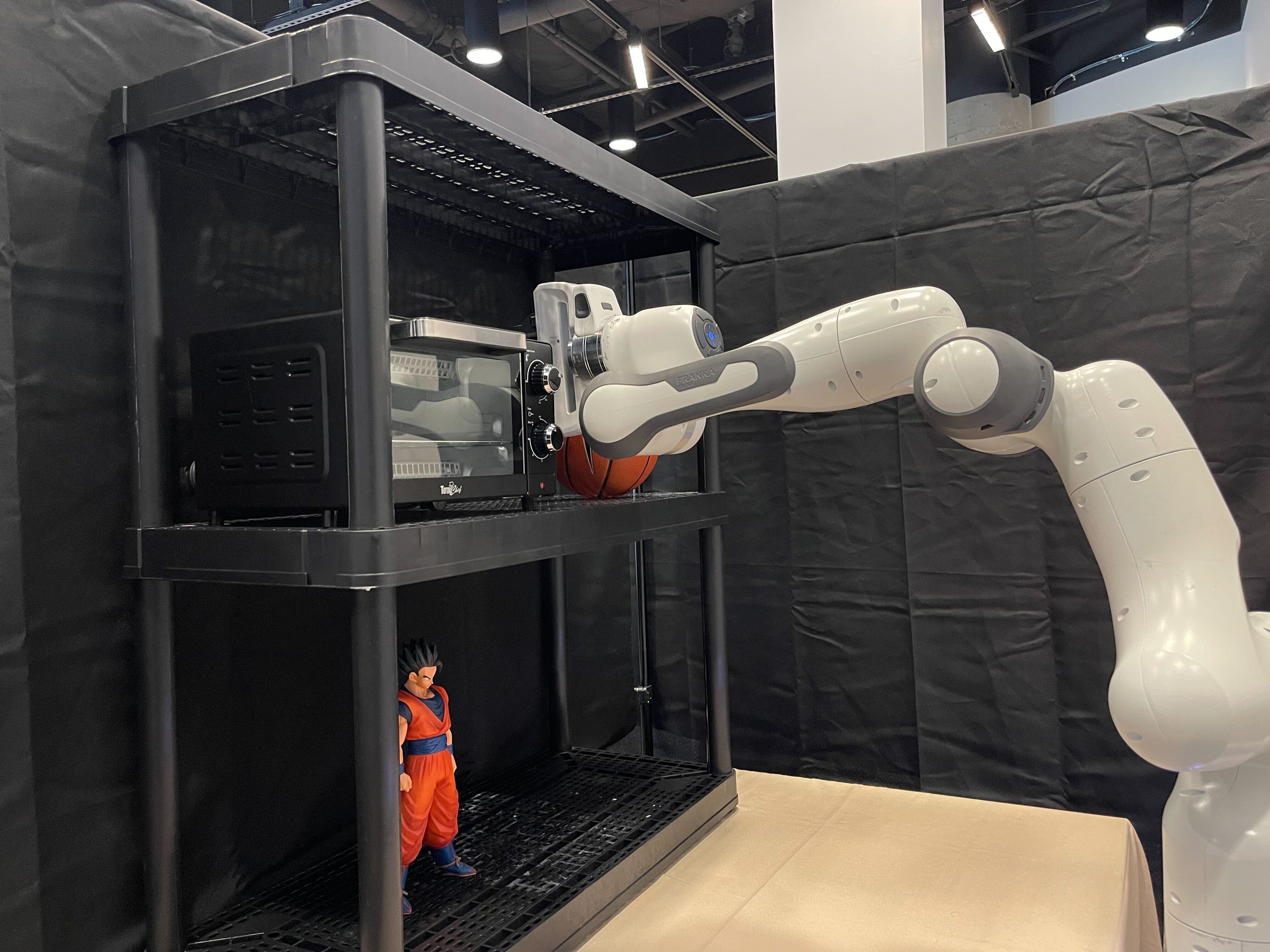}
    \caption{\small Shelf Scene 3}
\end{subfigure}
\begin{subfigure}[b]{0.24\linewidth}
    \includegraphics[width=\linewidth]{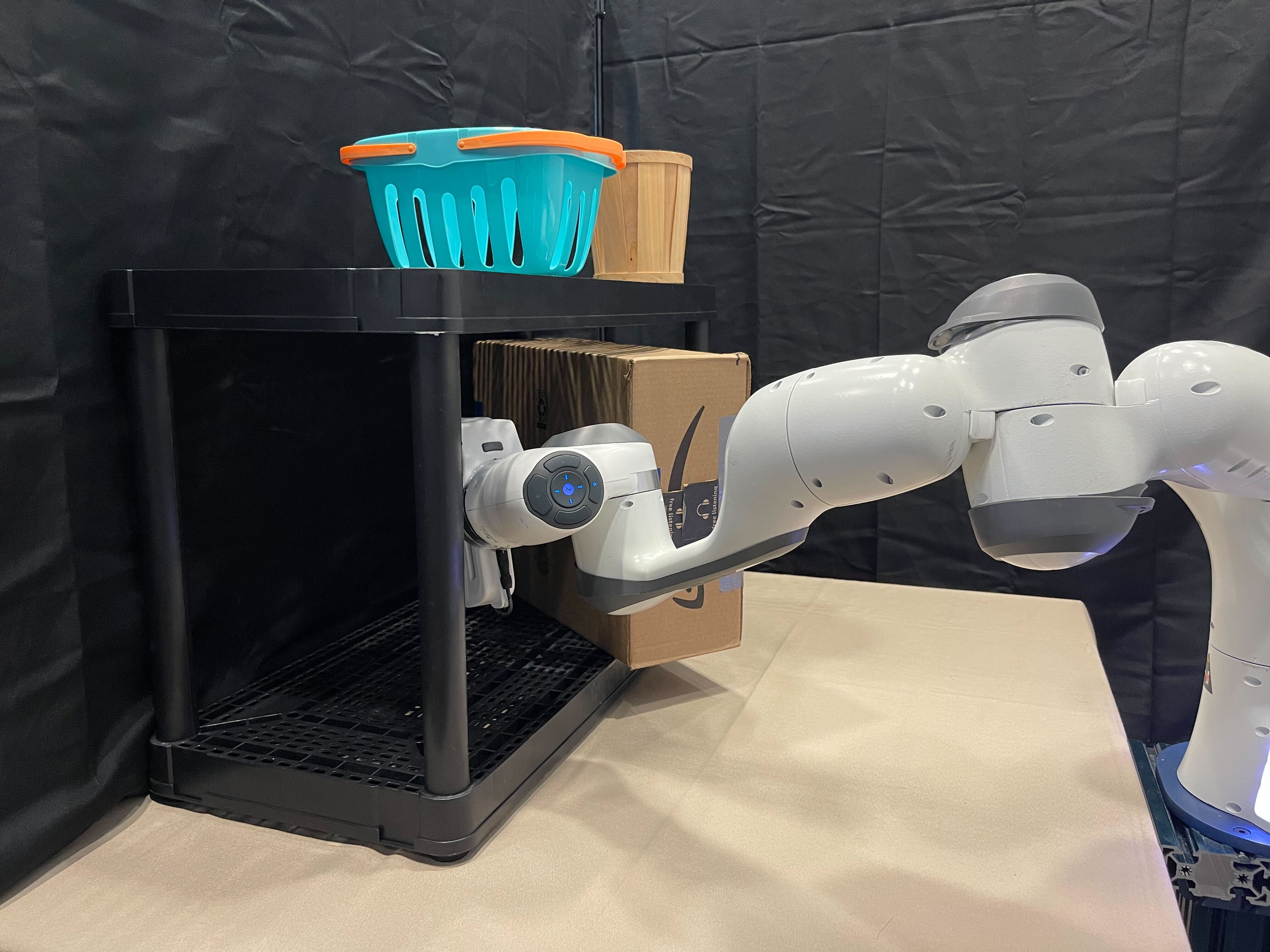}
    \caption{\small Shelf Scene 4}
\end{subfigure}
\begin{subfigure}[b]{0.24\linewidth}
    \includegraphics[width=\linewidth]{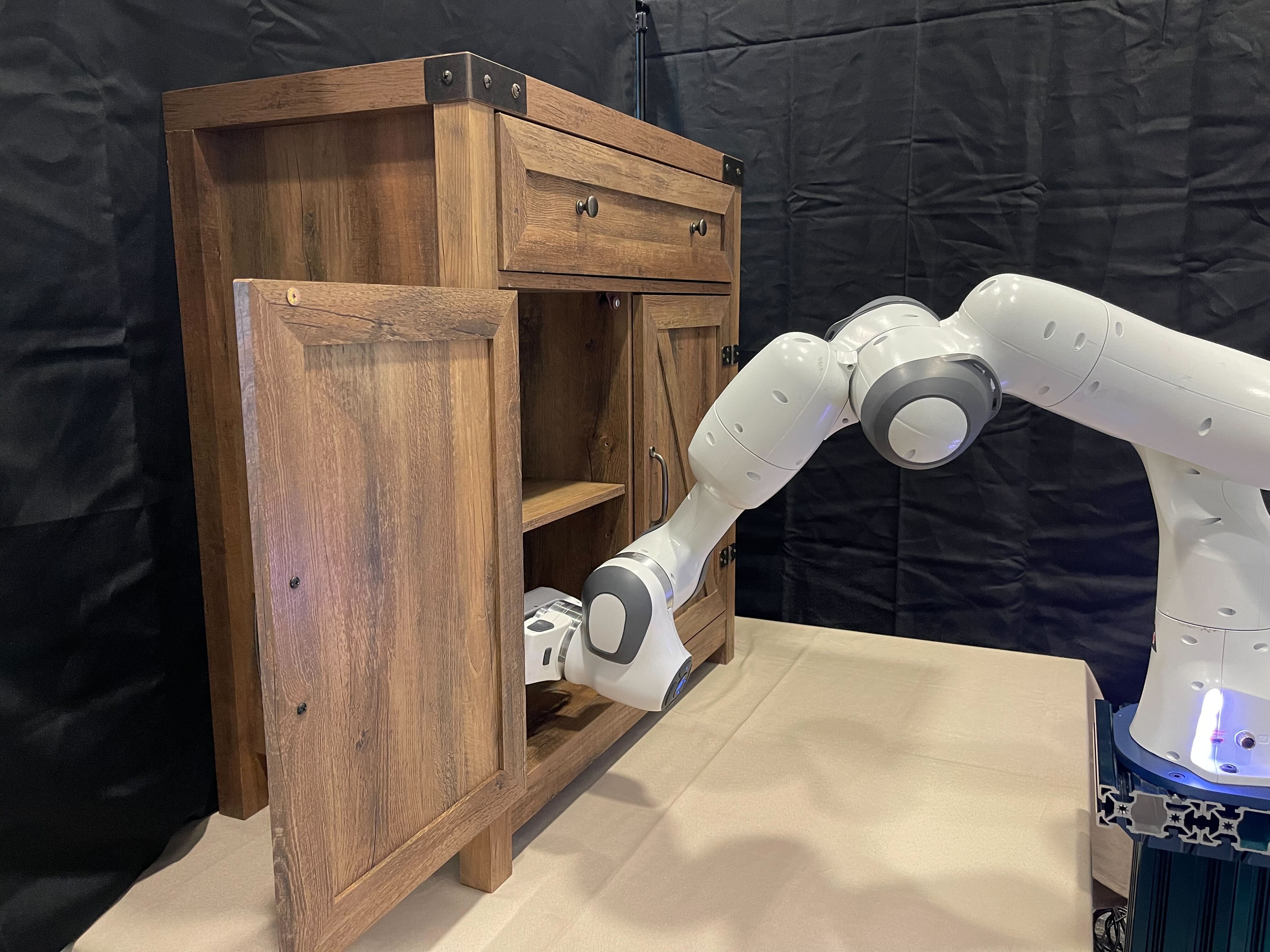}
    \caption{\small Articulated Scene 1}
\end{subfigure}
\begin{subfigure}[b]{0.24\linewidth}
    \includegraphics[width=\linewidth]{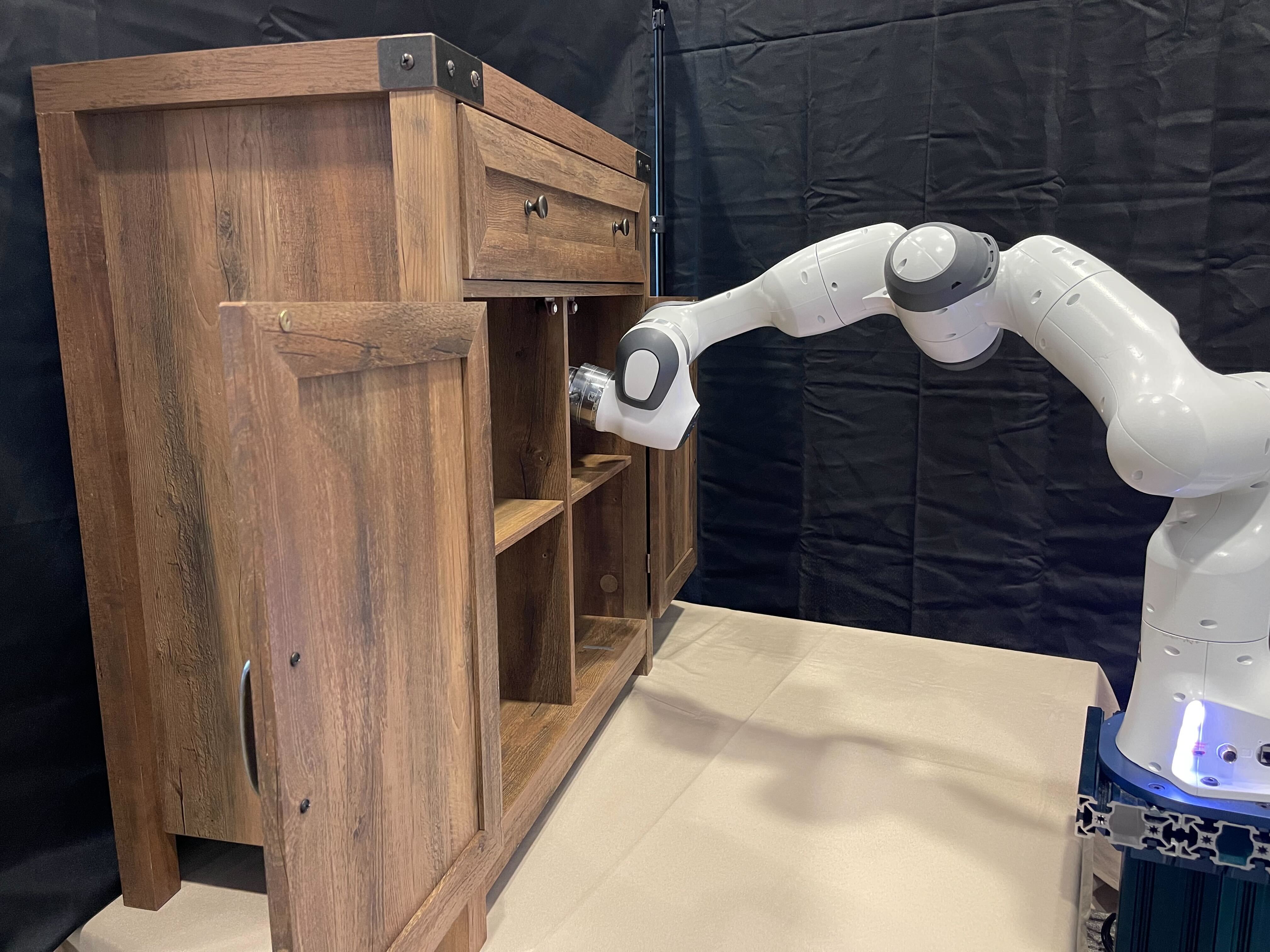}
    \caption{\small Articulated Scene 2}
\end{subfigure}
\begin{subfigure}[b]{0.24\linewidth}
    \includegraphics[width=\linewidth]{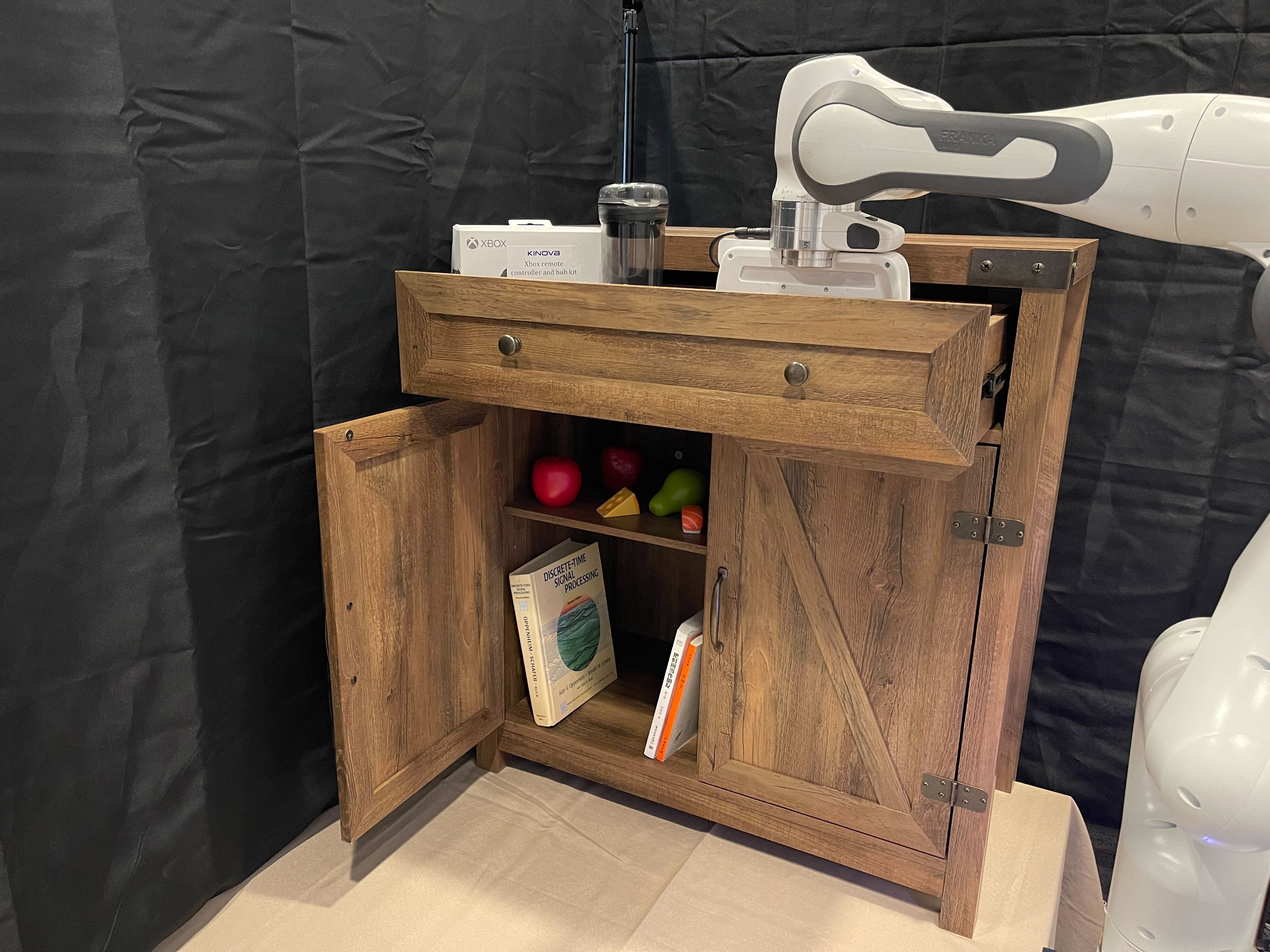}
    \caption{\small Articulated Scene 3}
\end{subfigure}
\begin{subfigure}[b]{0.24\linewidth}
    \includegraphics[width=\linewidth]{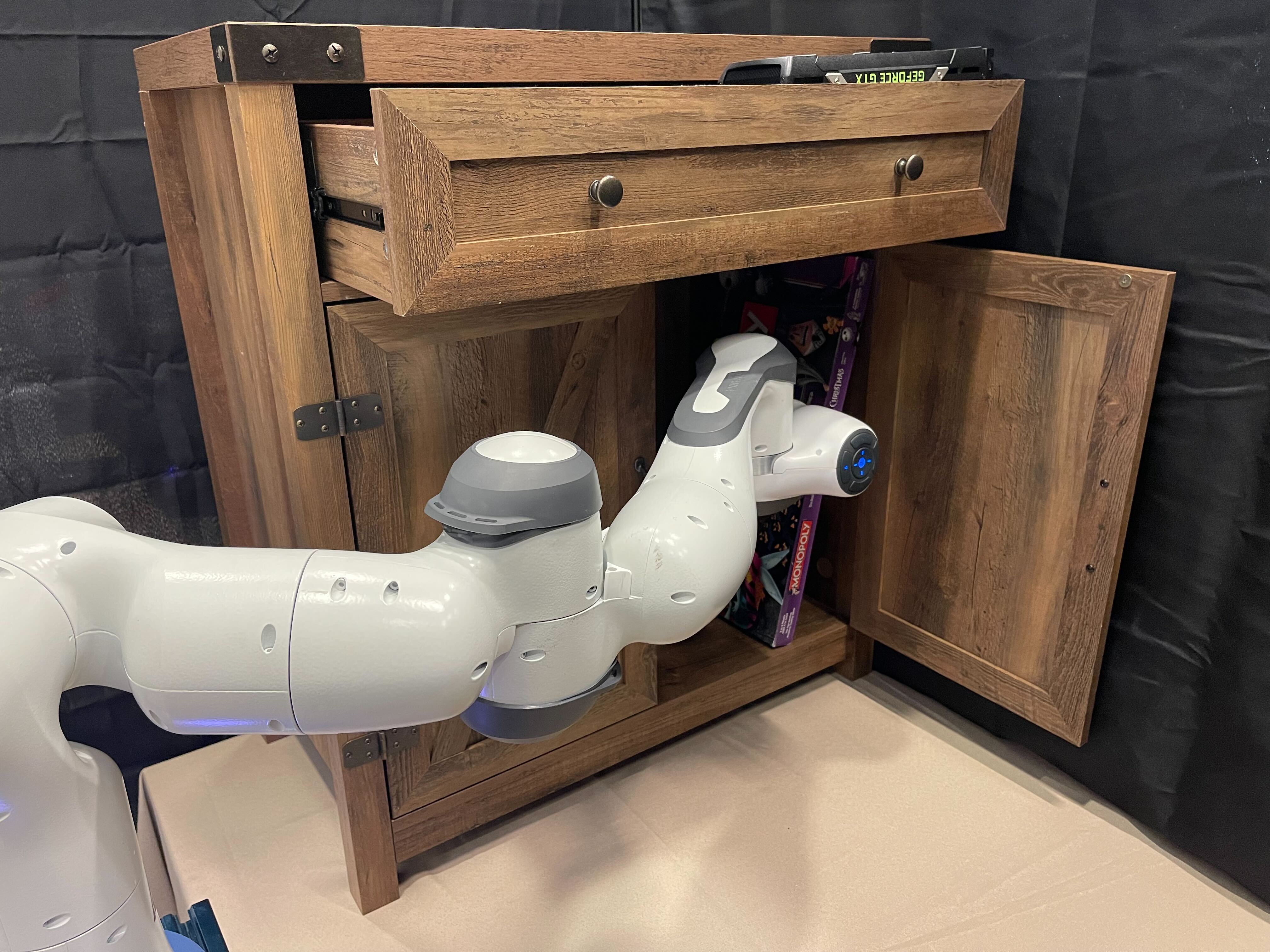}
    \caption{\small Articulated Scene 4}
\end{subfigure}
\begin{subfigure}[b]{0.24\linewidth}
    \includegraphics[width=\linewidth]{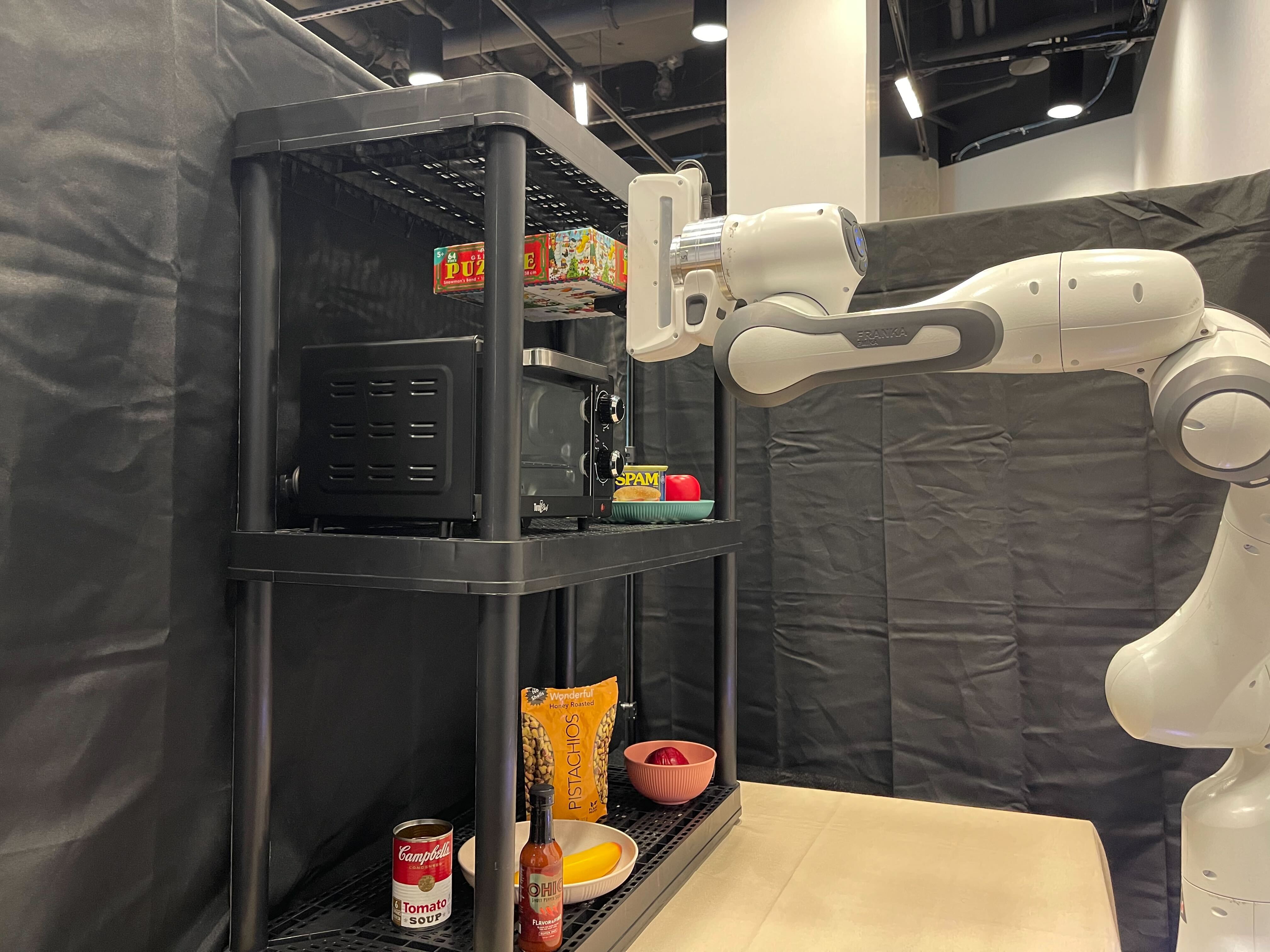}
    \caption{\small In Hand Object 1}
\end{subfigure}
\begin{subfigure}[b]{0.24\linewidth}
    \includegraphics[width=\linewidth]{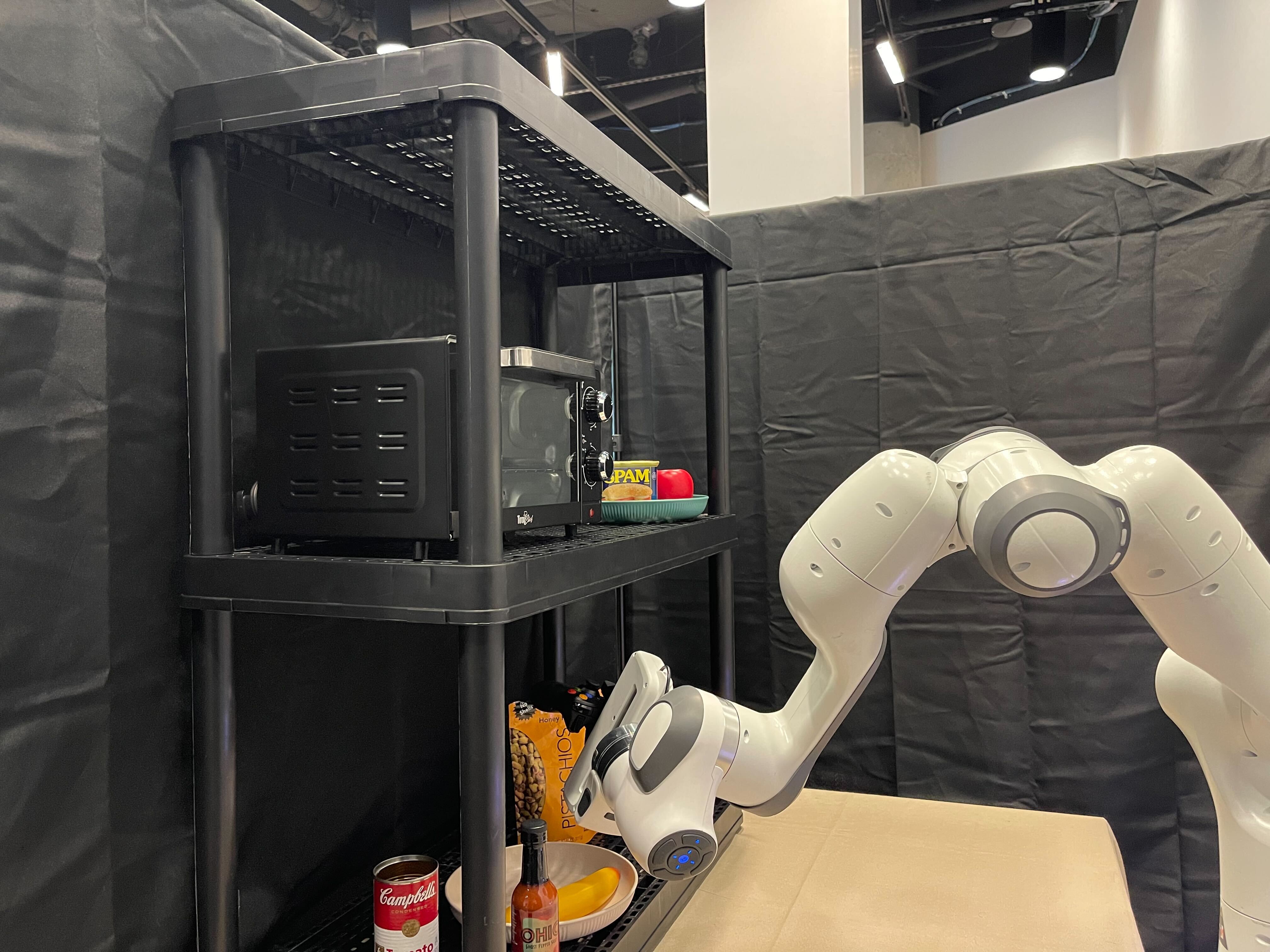}
    \caption{\small In Hand Object 2}
\end{subfigure}
\begin{subfigure}[b]{0.24\linewidth}
    \includegraphics[width=\linewidth]{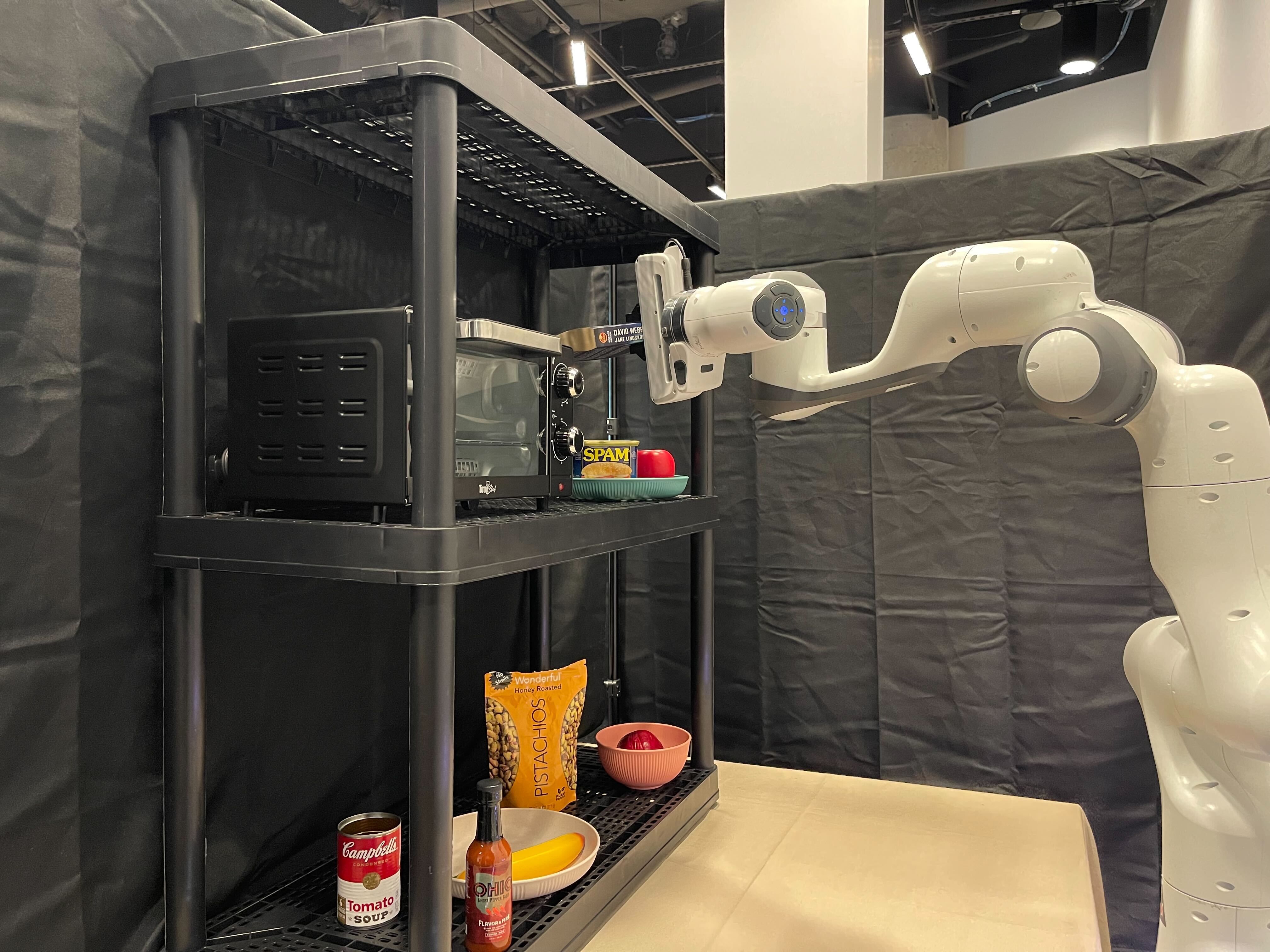}
    \caption{\small In Hand Object 3}
\end{subfigure}
\begin{subfigure}[b]{0.24\linewidth}
    \includegraphics[width=\linewidth]{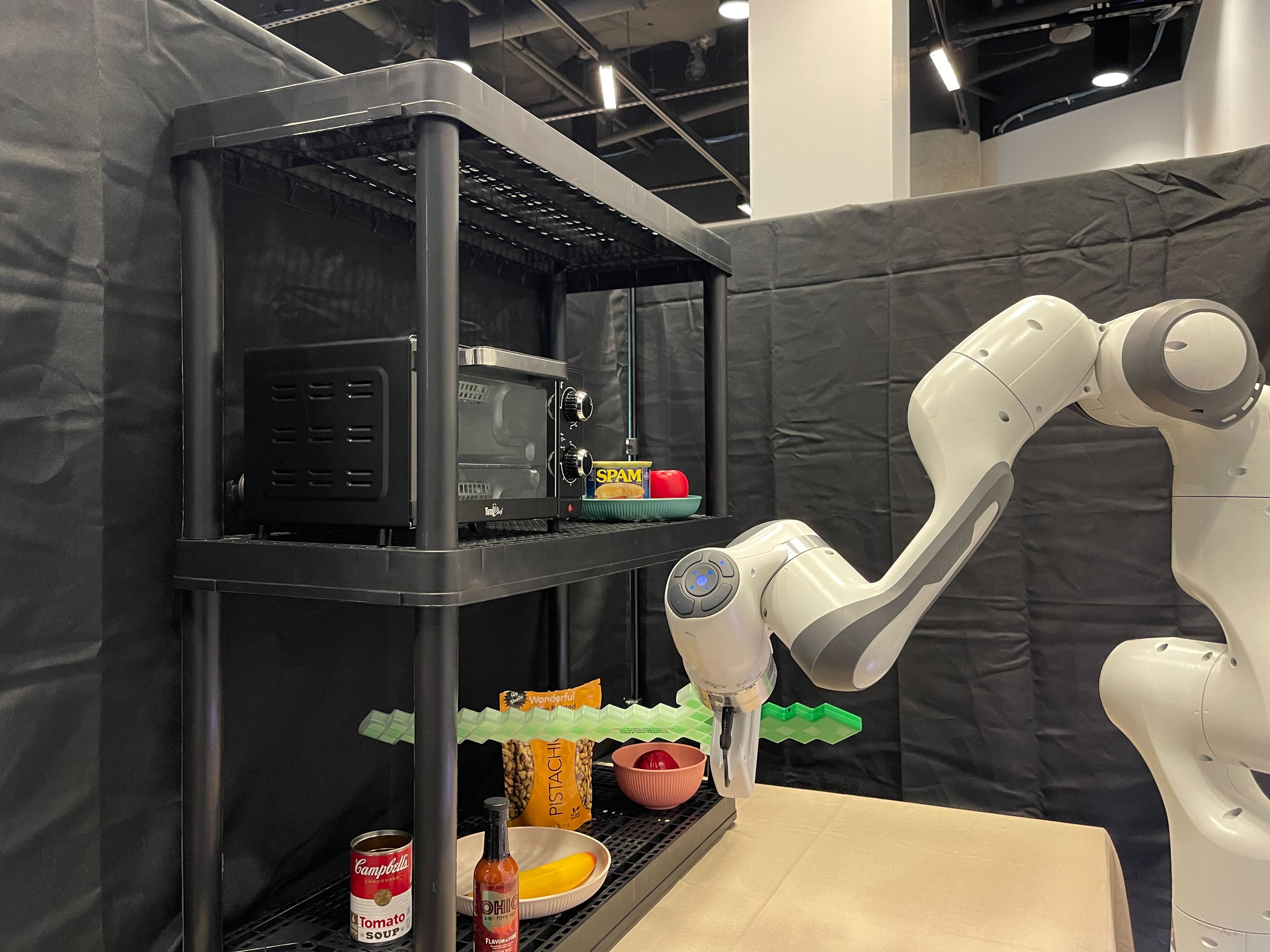}
    \caption{\small In Hand Object 4}
\end{subfigure}

\caption{\small Images of our 16 evaluation scenes.}
\vspace{-0.1in}
\label{fig:detailed setups}
\end{figure*}

\textbf{Bins}
This task requires the neural planner to perform collision avoidance when moving in-between, around and inside two different industrial bins pictured in the first row of Fig.~\ref{fig:detailed setups}. We randomize the position and orientation of the bins over the table and include the following objects as additional obstacles for the robot to avoid: toaster, doll, basketball, bin cap, and white box. The small bin is of size 70cm x 50cm x 25cm. The larger bin is of size 70cm x 50cm x 37cm. The bins are placed at two sides of the table. Between tasks, we randomize the orientation of the bins between 0 and 45 degrees and we swap the bin ordering (which bin is on the left vs. the right). The bins are placed 45cm in front of the robot, and shifted 60cm left/right.

\textbf{Shelf}
This task tests the agent's ability to handle horizontal obstacles (the rungs of the shelf) while maneuvering in tighter spaces (row two in Fig.~\ref{fig:detailed setups}). We randomize the size of the shelf (by changing the number of layers in the shelf from 3 to 2) as well as the position and orientation (anywhere at least .8m away from the robot) with 0 or 30 degrees orientation. The obstacles for this task include the toaster, basketball, baskets, an amazon box and an action figure which increase the difficulty. The shelf obstacle itself is of size 35cm x 80cm x 95cm. 

\textbf{Articulated}
We extend our evaluation to a more complex primary obstacle, the cabinet, which contains one drawer and two doors and tight internal spaces with small cubby holes (row three of Fig.~\ref{fig:detailed setups}). We randomize the position of the entire cabinet over the table, the joint positions of the drawer and doors and the sizes of the cubby holes. The obstacles for this task are xbox controller box, gpu, action figure, food toy, books and board game box. The size of the cabinet is 40cm x 75cm x 80cm. The size of the top drawer is 30cm x 65cm x 12cm. The size of the cubbies is 35cm x 35cm x 25cm. The drawer has an opening range of 0-30cm and the doors open between 0 and 180 degrees.  

\textbf{In-Hand Motion Planning}
In this task (shown in row four of Fig.~\ref{fig:detailed setups}), the planner needs to reason about collisions with not only the robot and the environment, but the held object too. We initialize the robot with an object grasped in-hand and run motion planning to reach a target configuration. For this task, we fix the obstacle (shelf) and its position (directly 80cm in front of the robot), instead randomizing across in-hand objects and configurations. We select four objects that vary significantly in size and shape: Xbox controller (18cm x 15cm x 8cm), book (17cm x 23cm x 5cm), toy sword (65cm x 10cm x 2cm), and board game (25cm x 25cm x 6cm). For this evaluation, we assume the object is already grasped by the robot, and the robot must just move with the object in-hand while maintaining its grasp. 

\subsection{Perception Visualization and Analysis}

\begin{figure}
    \vspace{-0.3in}
    \centering
    \includegraphics[width=0.45\linewidth]{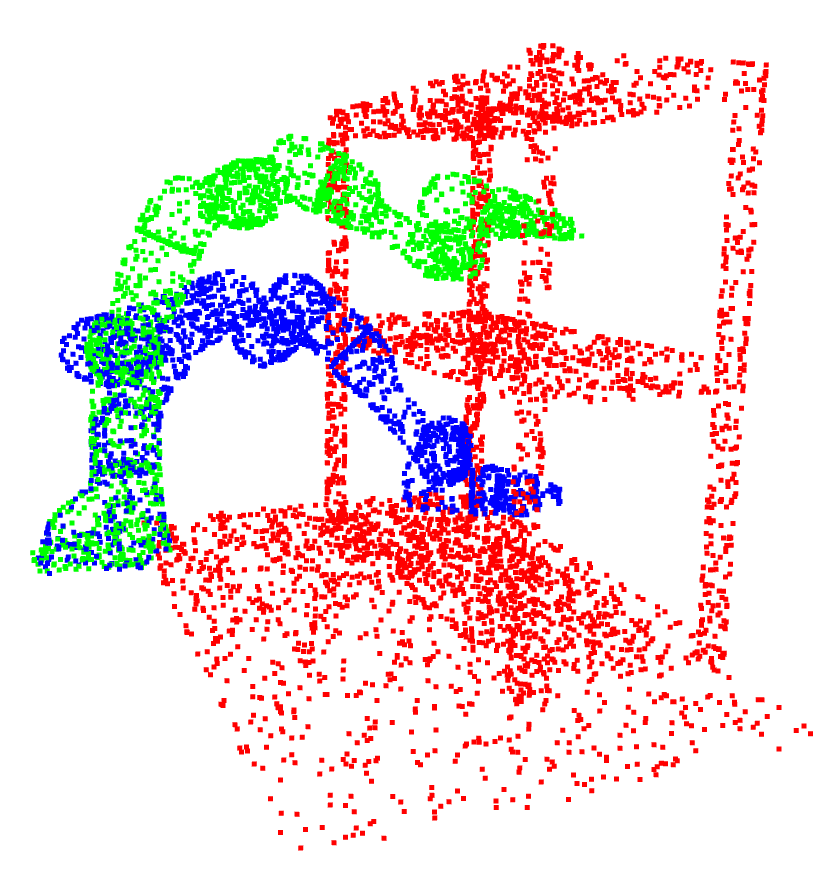} 
    \includegraphics[width=0.49\linewidth]{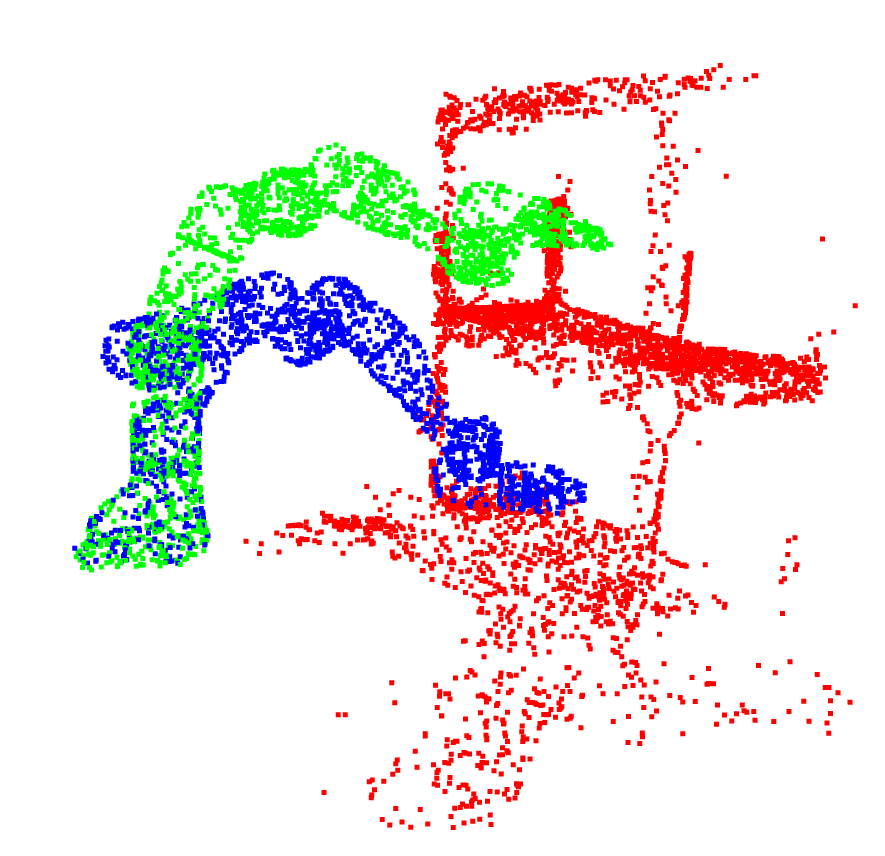} 
    \includegraphics[width=0.49\linewidth]{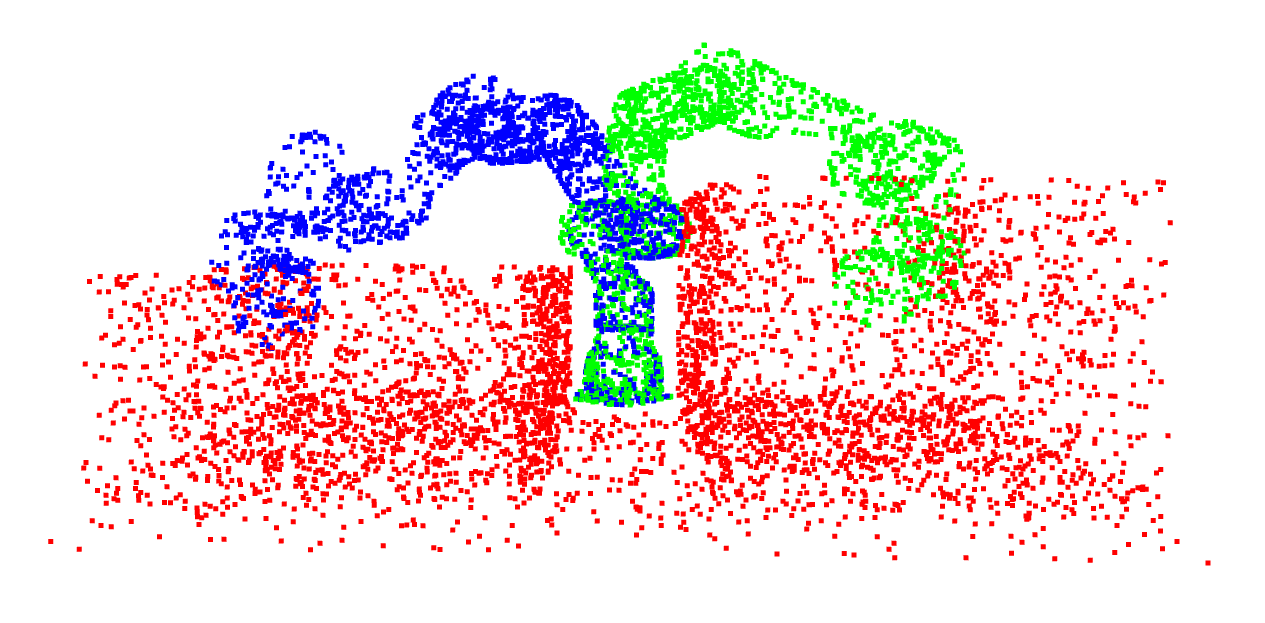}
    \includegraphics[width=0.49\linewidth]{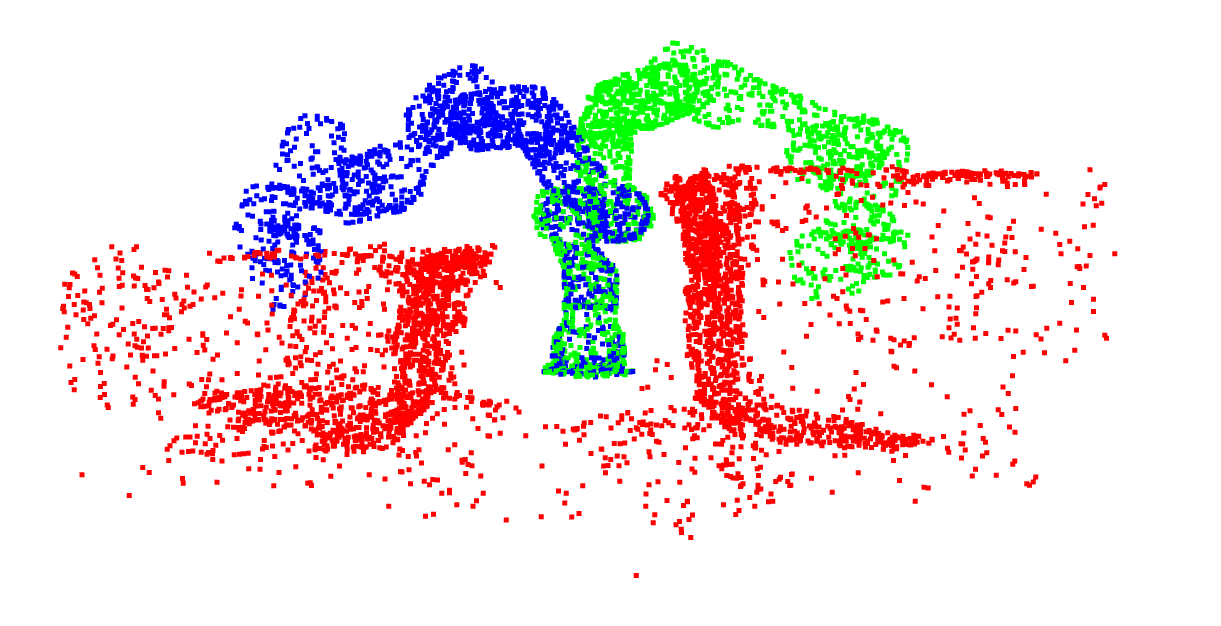}
    \caption{\small \textbf{Visualization of Sim and Real point-clouds}: We visualize point-clouds of the Bins and Shelf task in sim and real, in the same poses. Due to noise in depth sensing, the real world point-clouds have significantly more deformations, yet our policy generalizes well to these tasks.
    }
    \vspace{-0.12in}
    \label{fig:sim real point-cloud vis}
\end{figure}

We compare point-clouds from simulation and the real world for the Bins and Shelf task and analyze their properties. We replicate Bins Scene 4 and Shelf Scene 1 in simulation: simply measure the dimensions and positions of the real world objects and set those dimensions in simulation using the OpenBox and Shelf procedural assets. As seen in Fig.~\ref{fig:sim real point-cloud vis}, simulated point-clouds are far cleaner than those in the real world, which are noisy and perhaps more importantly, partial. The real-world point-clouds often have portions missing due to camera coverage as for large objects it is challenging to cover the scene well while remaining within the depth camera operating range. However, we find that our policy is still able to able operate well in these scenes, as PointNet++ is capable of handling partial point-clouds and is trained on a diverse dataset containing many variations of boxes and shelves with different types and number of components as well as sizes, which may enable the policy to generalize to partial boxes and shelves observed in the real world.

\end{document}